\documentclass[sigconf]{aamas}  

\usepackage{booktabs}
\usepackage{makecell,multirow,diagbox}
\usepackage{subfigure}
\usepackage{graphicx}
\usepackage{amsmath}
\usepackage{amssymb}
\usepackage{wrapfig}
\DeclareMathOperator*{\argmin}{arg\,min}

\setcopyright{ifaamas}  
\acmDOI{}  
\acmISBN{}  
\acmConference[AAMAS'18]{Proc.\@ of the 17th International Conference on Autonomous Agents and Multiagent Systems (AAMAS 2018)}{July 10--15, 2018}{Stockholm, Sweden}{M.~Dastani, G.~Sukthankar, E.~Andr\'{e}, S.~Koenig (eds.)}  
\acmYear{2018}  
\copyrightyear{2018}  
\acmPrice{}  

\begin{document}

\title{Efficient Reciprocal Collision Avoidance between Heterogeneous Agents Using CTMAT}  



\author{Yuexin Ma}
\affiliation{
  \institution{The University of Hong Kong}}
\email{yxma@cs.hku.hk}

\author{Dinesh Manocha}
\affiliation{
  \institution{University of North Carolina at Chapel Hill}}
\email{dm@cs.unc.edu}
\email{http://gamma.cs.unc.edu/CTMAT}

\author{Wenping Wang}
\affiliation{
  \institution{The University of Hong Kong}}
\email{wenping@cs.hku.hk}

\begin{abstract}  
We present a novel algorithm for reciprocal collision avoidance between heterogeneous agents of different shapes and sizes. We present a novel CTMAT representation based on medial axis transform to compute a tight fitting bounding shape for each agent. Each CTMAT is represented using tuples, which are composed of circular arcs and line segments. Based on the reciprocal velocity obstacle formulation, we reduce the problem to solving a low-dimensional linear programming between each pair of tuples belonging to adjacent agents. We precompute the Minkowski Sums of tuples to accelerate the runtime performance. Finally, we provide an efficient method to update the orientation of each agent in a local manner. We have implemented the algorithm and highlight its performance on benchmarks corresponding to road traffic scenarios and different vehicles. The overall runtime performance is comparable to prior multi-agent collision avoidance algorithms that use circular or elliptical agents. Our approach is less conservative and results in fewer false collisions.
\end{abstract}

%


\maketitle


\section{Introduction}

Computing collision-free trajectories for each agent is a fundamental problem in multi-agent navigation. The main goal is to ensure that each agent must take action to avoid collisions with other moving agents or obstacles and make progress towards its goal position. This problem has been well studied in AI and robotics~\cite{fraichard2004inevitable,bera2017sociosense,fox1997dynamic, godoy2016implicit}, VR~\cite{bruneau2015going,narang2016pedvr}, computer games and crowd simulation ~\cite{karamouzas2017implicit,stuvel2017torso,he2016dynamic,best2014densesense}, traffic simulation, emergent behaviors~\cite{reynolds1987flocks}, etc.

\begin{figure}
\subfigure[]{
\label{fig:trafficView_2}
\includegraphics[width=0.482\columnwidth]{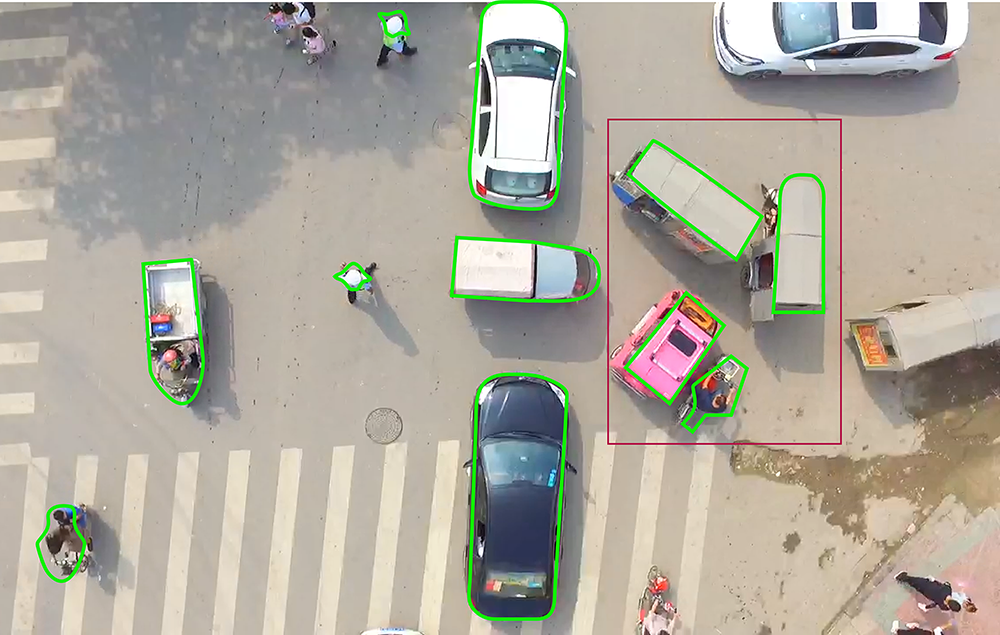}}
\subfigure[]{
\label{fig:trafficView_CTMAT}
\includegraphics[width=0.46\columnwidth]{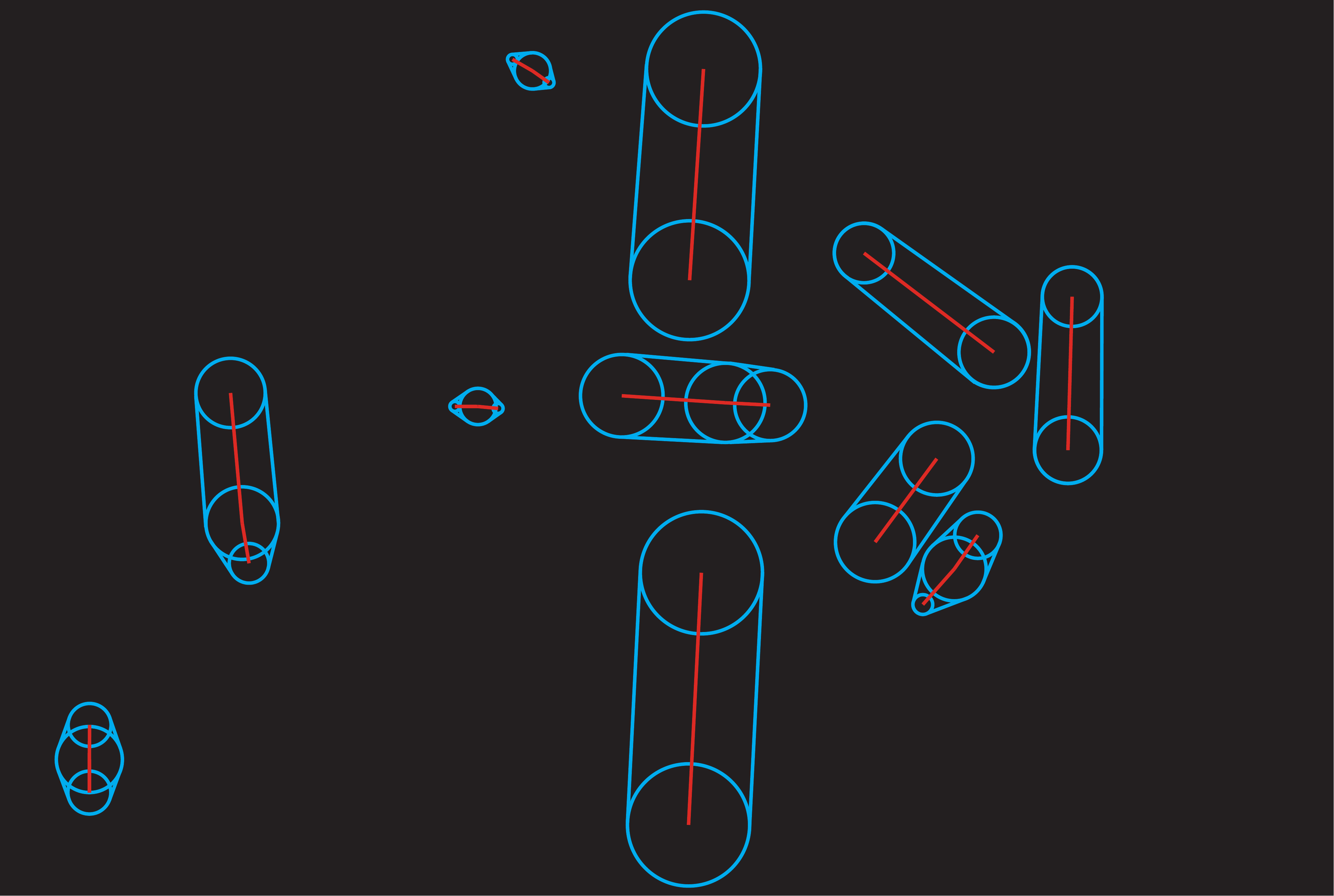}}

\subfigure[]{
\label{fig:compareGeometry}
\includegraphics[width=1\columnwidth]{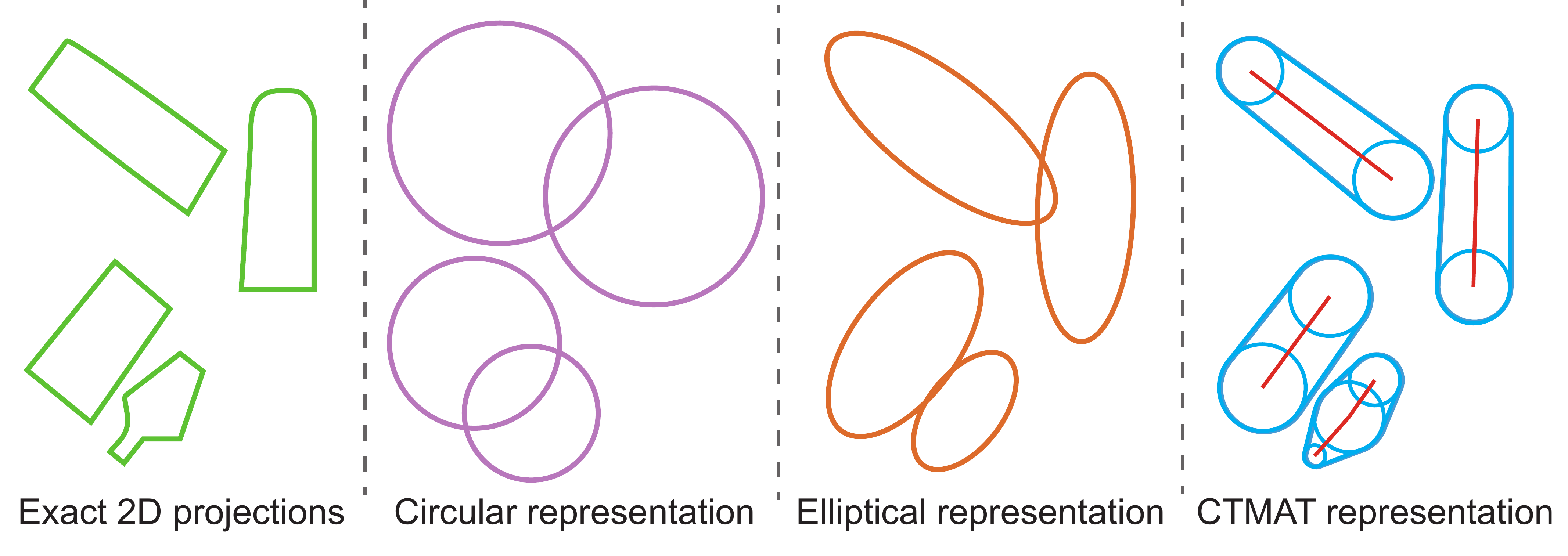}}
\caption{Traffic scenario. (a) Image of street traffic with different vehicles of varying sizes. The contours of vehicles are shown in green. (b) Our CTMAT representation for each agent with red medial axis. (c) Comparison of different kinds of representations of vehicle agents that lie in the red rectangle in (a). Our CTMAT representation is less conservative as compared to circles and ellipses and has similar runtime performance.}
\label{fig:traffic}
\vspace{-3ex}
\end{figure}

In this paper, we address the problem of computing collision-free paths for a large number of heterogeneous agents at interactive rates. Such agents are characterized by varying shapes and sizes (see Fig. \ref{fig:traffic}) and can be in close proximity.
In order to achieve real-time performance, most practical algorithms ~\cite{helbing1995social,reynolds1987flocks, van2011reciprocal, van2008reciprocal} use a  disc representation for each agent. However, one obvious disadvantage of using discs is that it results in a conservative approximation for each agent, especially when the shape is not round or has a large aspect ratio. This can result in a large number of false positives and the resulting multi-agent algorithms may not work well in dense scenarios. Other multi-agent algorithms use an ellipse~\cite{best2016real,narang2017interactive} or a capsule shape~\cite{stuvel2017torso} to represent each agent. However, these representations can become very conservative for some shapes and may also result in high number of false positives. The recent interest in autonomous driving simulators and navigation has motivated the development of a new set of multi-agent navigation algorithms for heterogeneous agents, whose shapes may correspond to cars, bi-cycles, buses, pedestrians, etc., that share the same road~\cite{best2017autonovi,pendleton2017perception, katrakazas2015real, saifuzzaman2014incorporating, ziegler2014making, sun2014trajectory, urmson2009autonomous, bareiss2015generalized}. Using a simple shape approximation like a circle or an ellipse for all agents can be very conservative for dense traffic situations, as shown in Fig. 1. Instead, we need efficient and accurate multi-agent navigation algorithms that can model agents with varying shapes and sizes.

\textbf{Main Results:} We present an algorithm for efficient collision-free navigation between heterogeneous agents by using a novel medial-axis agent representation (CTMAT). Based on CTMAT, we also present a reciprocal collision avoidance scheme (MATRVO). In order to represent  the heterogeneous agents for local navigation, we compute a compact representation of each agent based on the \textit{Medial Axis Transform} (MAT) ~\cite{biumtransformation}. Our formulation is based on a simplified discretization of the medial axis that captures the shape of agents. By linearly interpolating every two adjacent medial circles of MAT, we can get a set of tuples, which constitute CTMAT. Each tuple, composed of two circular arcs and two line segments (see Fig.~\ref{fig:exampleTuple}), is used to efficiently compute the Minkowski Sum, which is used for velocity obstacle computation for reciprocal collision avoidance. Our CTMAT representation can handle both convex and concave agents for reciprocal collision avoidance. We use precomputed tables of Minkowski Sums and precomputed width table of these agents to further accelerate the algorithm in handling a large environment that contains thousands of agents. We also update the orientation of each agent to generate collision-free trajectories. For the runtime performance, without precomputation, the average query time to perform the collision avoidance test between two CTMATs with one tuple is about $20$ microseconds on a single CPU core. In practice, our MATRVO algorithm with one tuple for each agent is $1.5-2$X slower than ORCA collision avoidance algorithms for circular agents, while it is $2$X faster than ERVO algorithm with elliptical agents.

The rest of the paper is organized as follows. We give a brief overview of prior work in collision avoidance and multi-agent navigation algorithms in Section 2. In Section 3, we provide an overview of MAT and how we construct our CTMAT representation for a given agent shape. In Section 4, we present our collision-free navigation algorithm (MATRVO) for heterogeneous agents. We present an acceleration scheme that uses precomputed Minkowski sums in Section 5 and present our approach to computing the orientation for local navigation in Section 6. We describe the implementation and highlight the results in Section 7. 
\begin{figure}
\includegraphics[width=0.9\columnwidth]{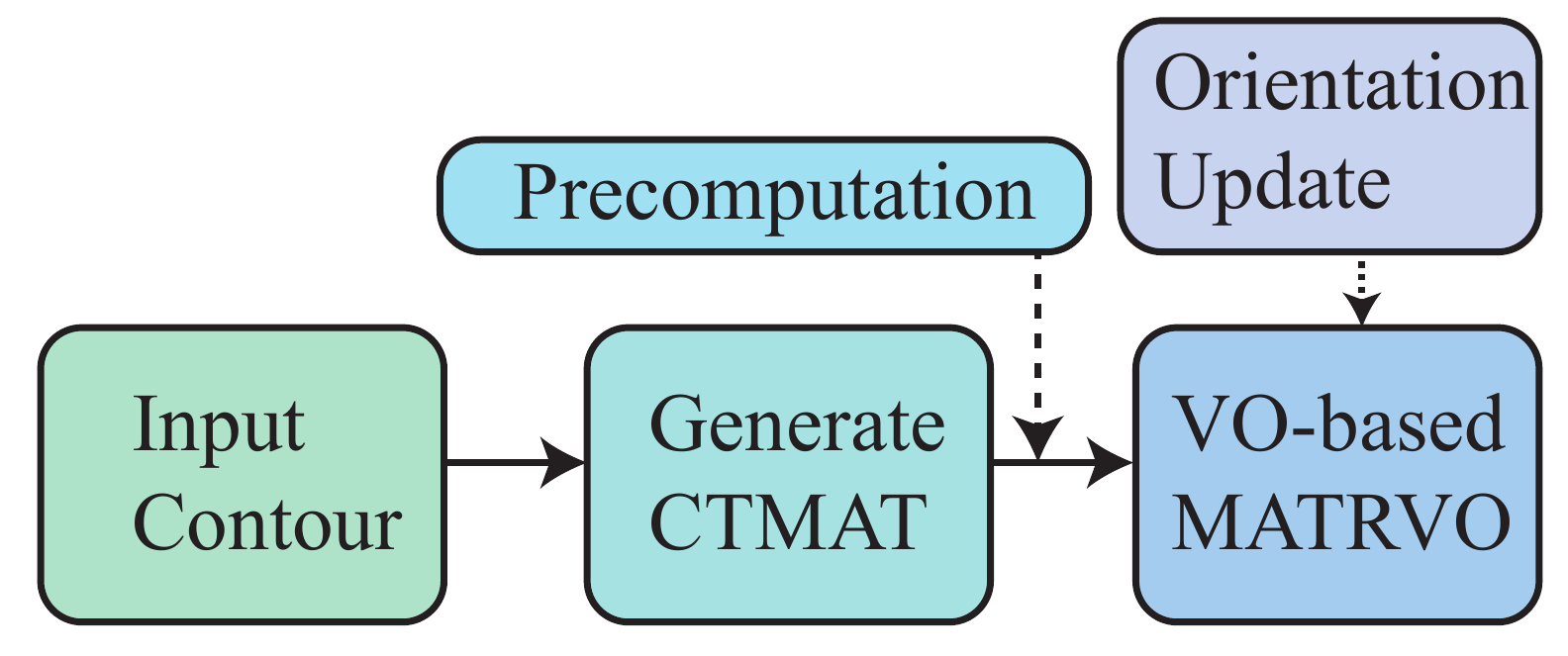}
\caption{Pipeline of our algorithm. For given contour of agent, we compute CTMAT representation. Then, we use MATRVO to compute velocities of agents. To speed up the algorithm, we can precompute Minkowski Sums and width table. We can update the orientation to match new velocities.}
\label{fig:Pipeline}
\vspace{-1ex}
\end{figure}

\section{Related Work}
The problem of collision avoidance and motion planning for robots or agents has been extensively studied. Many prior approaches~\cite{fraichard2004inevitable,borenstein1991vector,kanehiro2008local,simmons1996curvature} assume that the obstacles are static or slow moving as compared to the robot.
Prior algorithms for collision avoidance and collision-free navigation in dynamic environment can be classified into two categories, including centralized methods and decentralized methods. The former ~\cite{lavalle1998optimal,sanchez2002using,vsvestka1998coordinated} regard all the agents as part of a single global system and decide the actions for each agent in the unified configuration space. These methods can provide global guarantees, but their complexity increases significantly with the number of agents in the scene. In practice, they are limited to scenarios with a few agents. Decentralized methods ~\cite{fox1997dynamic, hsu2002randomized, sanchez2002using, petti2005safe, karamouzas2017implicit, godoy2016implicit} compute the motions and trajectories for the agents independently. They usually make use of local environmental information to compute a local trajectory according to agents' positions and current motions.  Some of these earlier methods did not account for reactive behaviors of other agents. 

Among decentralized approaches, velocity obstacle (VO)~\cite{fiorini1998motion} is a widely used algorithm for collision avoidance for a robot navigating among dynamic obstacles. It has been extended to model reciprocal behaviors between agents~\cite{van2008reciprocal,van2011reciprocal} and can provide sufficient conditions for collision avoidance. Furthermore, they can also account for kinematic and dynamic constraints of agents~\cite{bareiss2013reciprocal, alonso2012reciprocal, lalish2012distributed}. All these methods assume that each agent is represented as a disc. Some other local navigation methods for disc-based agents include cellular decomposition~\cite{schadschneider2001cellular}, rule-based methods ~\cite{reynolds1987flocks} and force-based methods ~\cite{helbing1995social,karamouzas2009predictive,pelechano2007controlling}.

In order to better approximate human and robot shapes, efficient reciprocal velocity obstacle methods have been proposed for elliptical agents~\cite{best2016real}. The resulting algorithm extends ORCA and takes advantage of the precomputed table of Minkowski Sums of ellipses to compute velocities for all the agents in real time. Based on this ellipse representation, Narang  et al.~\cite{narang2017interactive} can accurately model human motions in dense situations, including shoulder turning and side-stepping. A capsule-shaped approximation of agents has been used for character animation~\cite{stuvel2017torso} to generate torso-twisting and side-stepping characters of crowd model. All these methods are less conservative than disc-based agent representations and can also model orientation changes. However, they are limited to homogeneous environments, where each agent has the same shape.  It could be quite expensive to extend these methods to heterogeneous environments where the agents have different sizes and shapes.

\begin{figure}
\subfigure[]{
\label{fig:exampleMA_1}
\includegraphics[width=0.26\columnwidth]{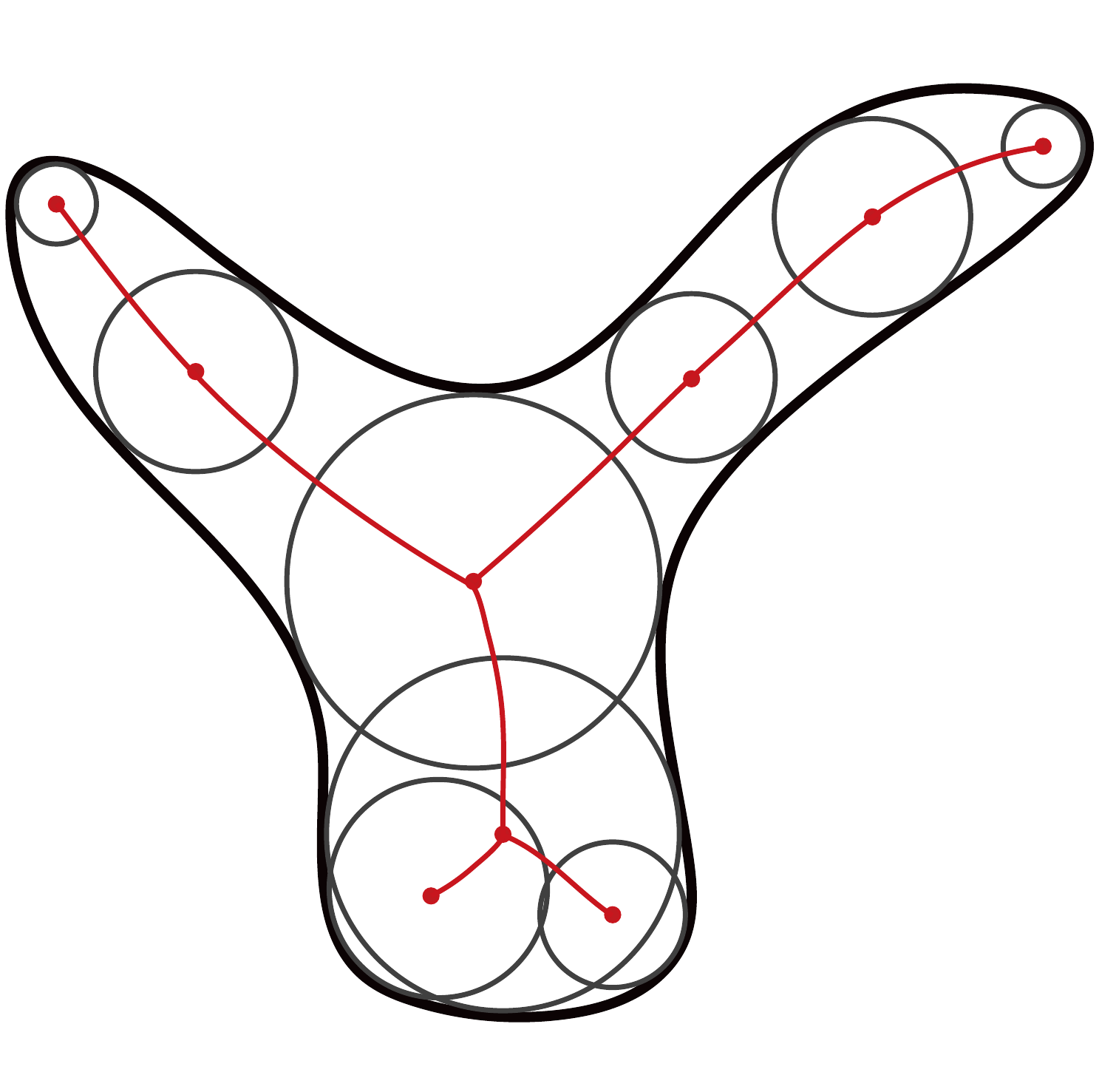}}
\subfigure[]{
\label{fig:exampleTuple}
\includegraphics[width=0.14\columnwidth]{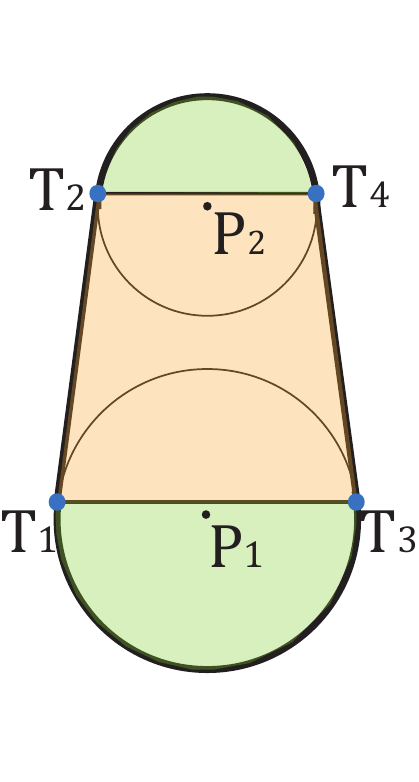}}
\subfigure[]{
\label{fig:exampleMA_3}
\includegraphics[width=0.26\columnwidth]{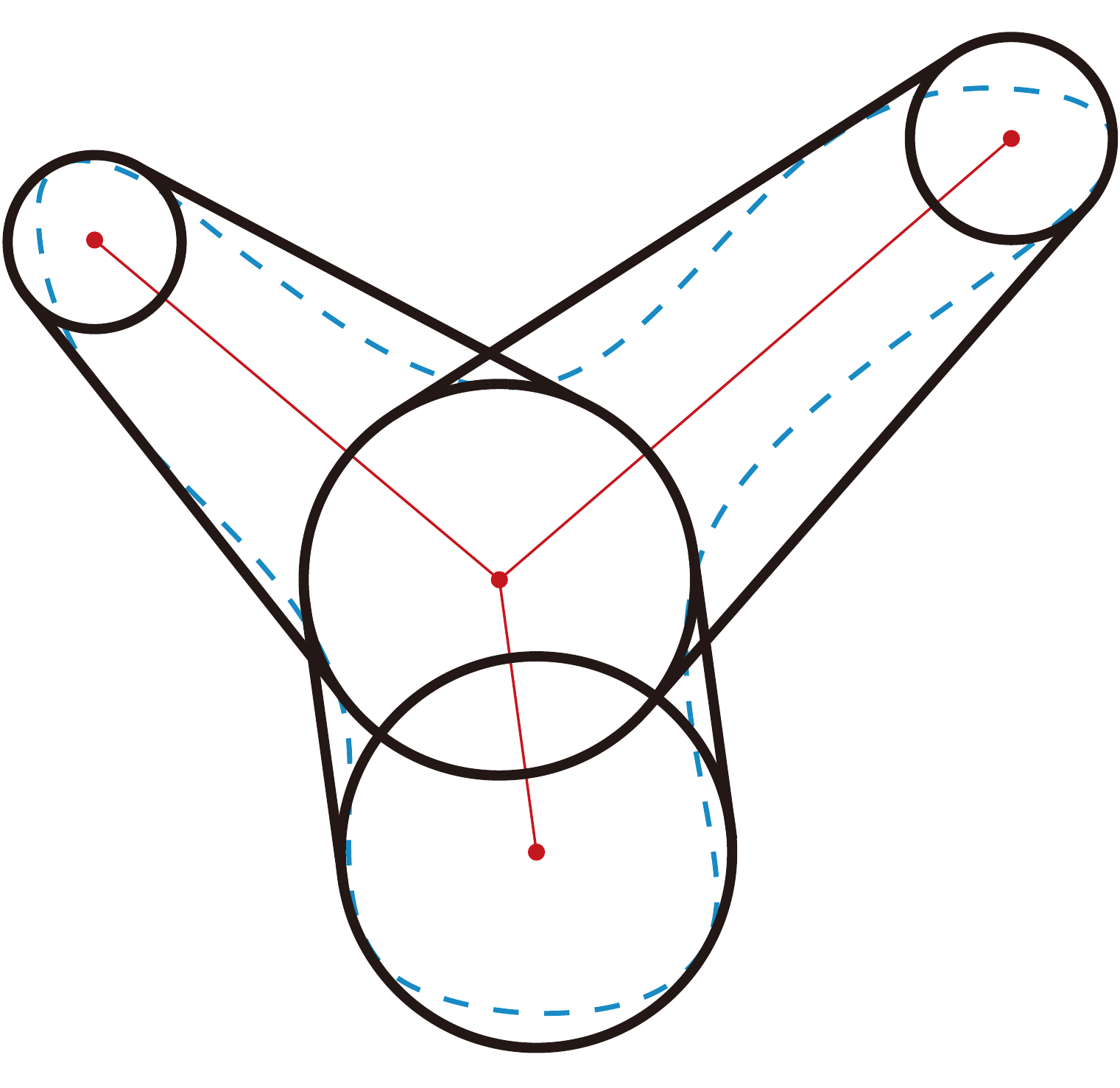}}
\subfigure[]{
\label{fig:exampleMA_4}
\includegraphics[width=0.26\columnwidth]{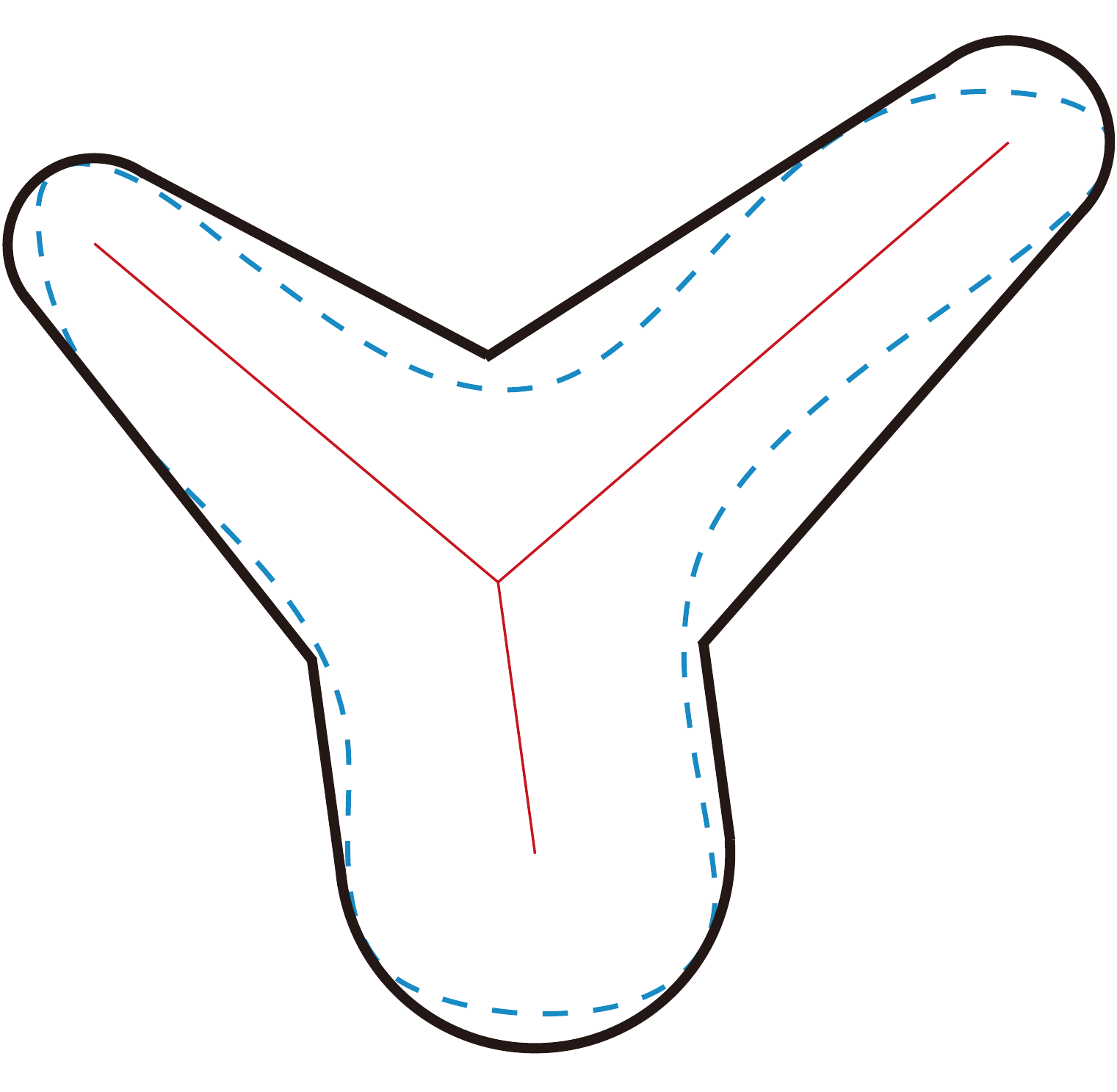}}
\caption{(a) MAT of a 2D shape: Red curve is the medial axis and black circles are several sampling medial circles of the shape. (b) Tuple: Linear interpolation of two neighboring medial circles. (c) Simplified combination of tuples by our algorithm. Blue dotted curve is the contour of the original shape. (d) CTMAT representation of the original shape.}
\label{fig:MA}
\vspace{-3ex}
\end{figure}

\begin{figure*}
\vspace{-2ex}
\subfigure[]{
\label{fig:CTMAT_1}
\includegraphics[width=0.415\columnwidth]{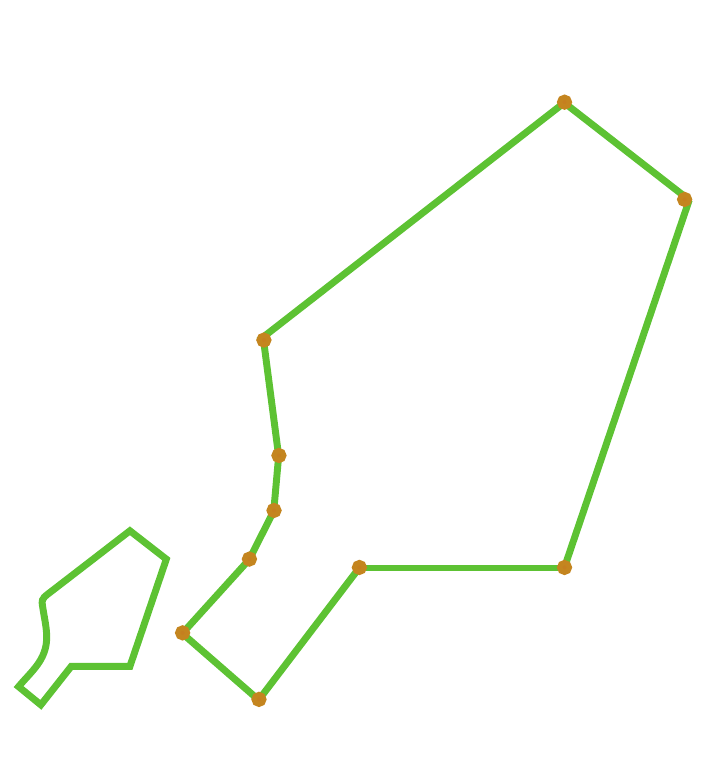}}
\subfigure[]{
\label{fig:CTMAT_2}
\includegraphics[width=0.3\columnwidth]{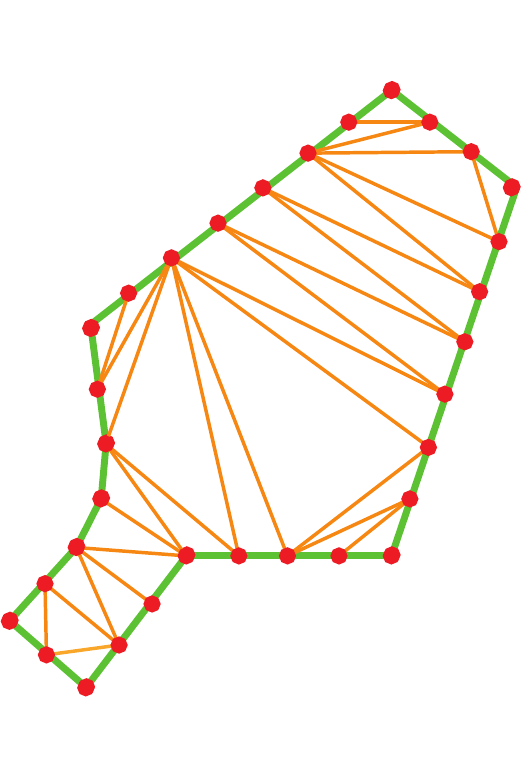}}
\subfigure[]{
\label{fig:CTMAT_3}
\includegraphics[width=0.315\columnwidth]{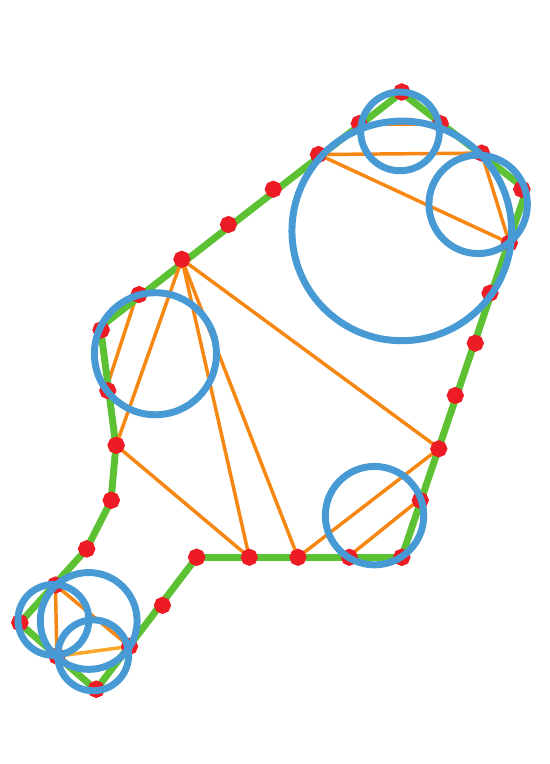}}
\subfigure[]{
\label{fig:CTMAT_4}
\includegraphics[width=0.3\columnwidth]{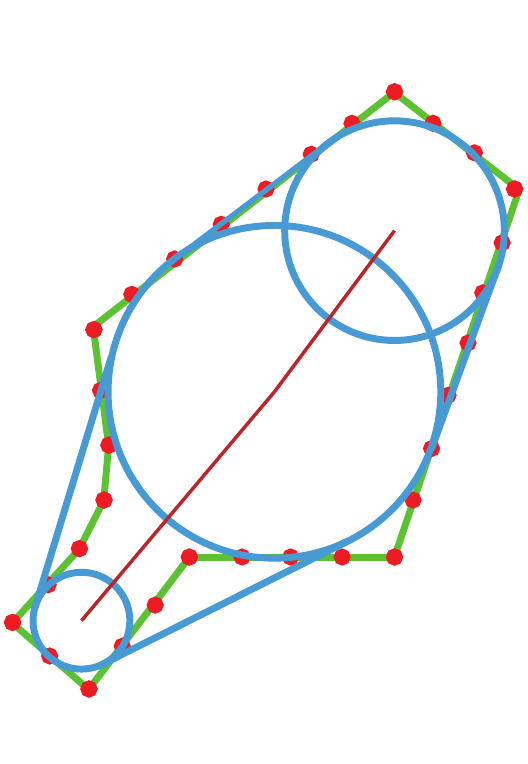}}
\subfigure[]{
\label{fig:CTMAT_5}
\includegraphics[width=0.31\columnwidth]{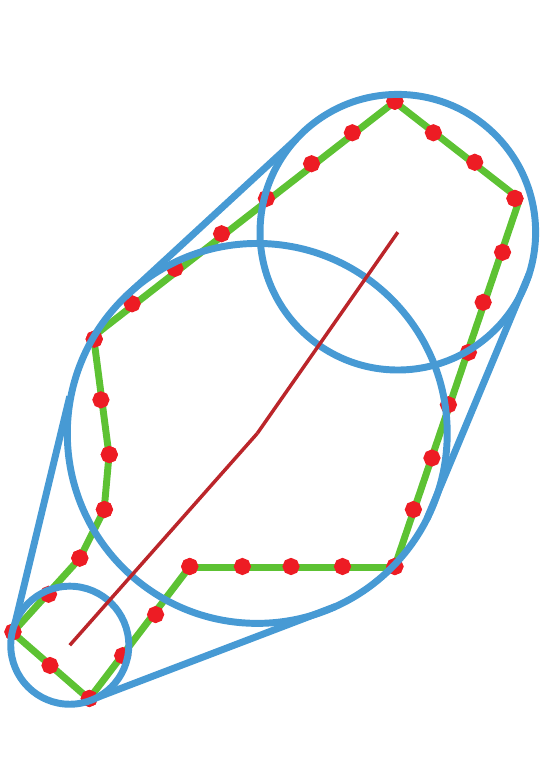}}
\subfigure[]{
\label{fig:CTMAT_6}
\includegraphics[width=0.31\columnwidth]{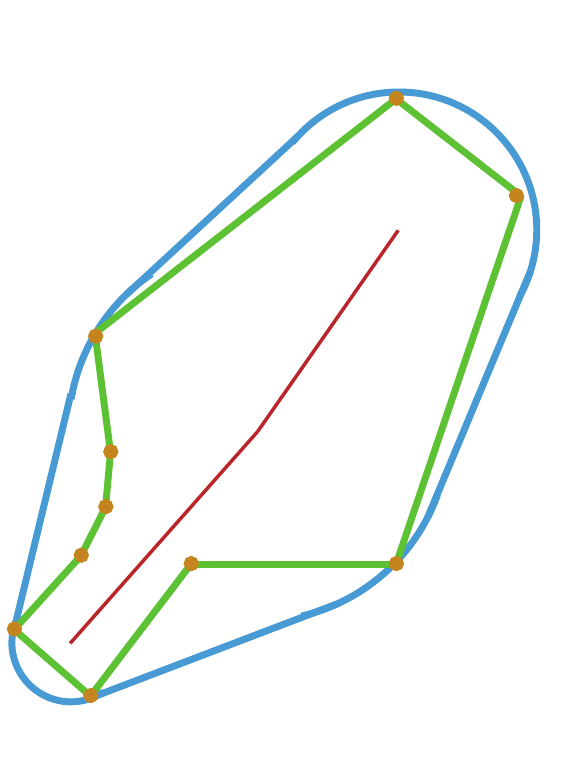}}
\caption{The process of generating CTMAT for a given agent: (a) Transfer the curved contour of agent to a polygon. (b) Sample on the boundary of polygon and compute the CDT of the point samples. (c) Compute circumcircles for J-triangles and T-triangles. (d) Select valid circles as medial circles and compute  the corresponding tuples. (e) Modify tuples to cover the agent. (f) The CTMAT.}
\label{fig:CTMAT}
\end{figure*}

\section{CTMAT: Approximation of Agents}
Our approach is designed for heterogeneous environments, where each agent could have a different shape that is convex or non-convex. Our algorithm only assumes that we are given the 2D boundary or contour of each agent. One of our goals is to compute a representation that is tight--fitting and useful for efficient reciprocal collision avoidance for a large number of agents. We present a new representation that exploits the properties of the medial axis transform of the object. 

\subsection{Medial Axis Transform}
The medial axis, proposed by \cite{biumtransformation}, is an intrinsic shape representation that naturally captures the symmetry and interior properties of an object. In 2D, the medial axis of a shape, which is bounded by planar curve $C$, is the locus of the centers of circles that are tangent to curve $C$ at two or more points, as shown in Fig. ~\ref{fig:exampleMA_1} shows. The circles are called medial circles. If the distance to the boundary is regarded as the radius of a medial axis point, we obtain the \textit{Medial Axis Transform} (MAT), denoted as $(P, r)\in R^d$ , $d = 2$ or 3, where $P$ is the center of the medial circle and $r$ is the radius. The contour of an object can be reconstructed from MAT and the accuracy of the reconstructed shape is related to the number of sampled medial points.

\subsection{CTMAT Representation}
The exact representation of a medial axis is a continuous shape, but it is hard to compute. In most cases, we just compute a discretized representation in terms of a set of medial circles and their neighboring relationship, which is defined by the relative locations of the centers on the medial axis. When reconstructing a shape from the stored discrete information, we perform linear interpolation~\cite{li2015q} between any pair of neighboring medial circles. Finally, the outermost contour is the reconstructed shape approximation of the agent. Fig. \ref{fig:exampleTuple} shows a linear interpolation of two neighboring sampling medial circles. We call its contour a tuple, which is a basic unit of our representation. The tuple consists of two line segments $T_1T_2$, $T_3T_4$ and two circular arcs $Arc(T_1T_2)$, $Arc(T_3T_4)$ of the green area. After interpolation, the contour of the set of tuples of the original object corresponds to our CTMAT representation. The CTMAT representation of an agent $A$ is denoted as $CTMAT(A)$, which is also composed of line segments and circular arcs. To improve the efficiency of our algorithm, we make a tradeoff between the number of tuples and the conservative nature of our representation (Fig. \ref{fig:exampleMA_3}) to approximate the original shape. The final CTMAT is shown in Fig. \ref{fig:exampleMA_4}. In order to illustrate the structure of CTMAT, we the draw detailed combination of tuples with medial axis inside in the figure along with the contour to represent CTMAT.

\subsection{CTMAT Computation}
Our approach to computing the CTMAT consists of two parts - computing the reconstructed shape of the simplified medial axis and modifying the tuples to enclose the agent. Fig. \ref{fig:CTMAT} shows our pipeline.
To ensure that the contour of agent can be easily represented, we first approximate the closed curve with a polygon, which is similar to the original shape and overestimates it. Many methods have been  proposed~\cite{amenta2001power, lee1982medial,culver2004exact} to compute the MAT.  Among them, Voronoi-based approaches ~\cite{brandt1992continuous, attali1997computing,attali2001delaunay,dey2004approximate} are widely used in practice. In 2D, the Voronoi vertices approximate the medial axis and converge to the exact medial axis if the sample density of the points on the contour approaches infinity. Therefore, we uniformly sample on the polygonal contour. All the vertices of the polygon are added to the samples. The sampling density depends on user-specified accuracy. Given a set of sampling points of agent's boundary, one easy method to compute Voronoi vertices is to compute Delaunay Triangulation (DT)~\cite{watson1981computing}. For convex shapes, we just compute DT. For non-convex shapes, we use Constraint Delaunay Triangulation (CDT)~\cite{chew1989constrained} to guarantee that all the sampling line segments, which is constituted by two adjacent sampling points, belong to the edges of triangulation. Moreover, we need to delete the outer triangles. Fig. \ref{fig:CTMAT_2} shows the result of this step for a non-convex shape. The complexity of computing DT or CDT is $O(n\log n)$, where $n$ is the number of sample points. There are three kinds of triangles~\cite{prasad1997morphological} in the resulting triangulation. Those triangles with two external edges located in the terminal of a branch or a protrusion of the shape are called T-triangles. Those triangles with one external edge are S-triangles. Finally, the triangle with no external edges determines a junction of branches of the polygon is the J-triangle. The circumcircle of each triangle can be used to approximate the medial circle. The neighboring relationship is decided by their related triangles. If two triangles share the same edge, they are neighbors of each other. However, using a CTMAT representation with too many medial circles can slow down the runtime collision avoidance scheme. It turns out that performing the interpolation based on the T-triangle and J-triangle would cover the S-triangles, and we only compute the medial circles based on T-triangles and J-triangles, as shown in Fig. \ref{fig:CTMAT_3}. If there is a set of consecutive S-triangles between two circles, we think these two circles are adjacent. To further simplify the representation, we use the following conditions to select the final valid medial circles and compute corresponding tuples for CTMAT.
 \begin{equation}
    {\Gamma_{ij}} = \frac{d(c_i, c_j)+r_j}{r_i+r_j},
  \end{equation}
where $c_i$, $ c_j$ and $r_i$, $ r_j$ are the centers and the radii of two neighboring medial circles respectively, and $d(c_i, c_j)$ represents the distance between the two centers. If $\Gamma_{ij} < \varphi$, we delete one of the two medial circles. If the medial circles of two related triangles are of different types, we delete the circle corresponding to the T-triangle. Otherwise, we remove the smaller one. After deleting a circle, the neighboring relationships of the removed circle are added to the remaining one. $\varphi$, standing for the threshold of filter, is a user-defined parameter and we use $\varphi = 1$ in our experiment. Fig. \ref{fig:CTMAT_4} shows the result after filtering.

After computing the simplified combination of tuples, we modify that representation so that it totally covers the agent. We use an optimization algorithm for this modification step with the goal of generating as small tuples as possible to contain the boundary of the agent and maintaining the original tuples' shape (Fig. \ref{fig:CTMAT_4}) as much as possible. Assume the variables of the current medial circle $m$ are represented as the center $c$ and the radius $r$. The optimization algorithm is applied to each medial circle and can be expressed as:
\begin{align*}
\min\,\,  &E= E_1 + E_2, \\ 
&E_1 = d^2(c, c_o),\\
&E_2 =\left\{
\begin{aligned}
0\hspace{1cm}&, r<r_o \\
(r-r_o)^2&, otherwise
\end{aligned}
\right.\\
\tag{2}
&\texttt{s}.\texttt{t}.\quad
\begin{cases}
G_{in} &\in U\\
G_{out} &\in U,
\end{cases}
\end{align*}
where $c_o$ and $r_o$ are the original center and radius of $m$, $E_1$ stands for the distance from the initial position and $E_2$ stands for the difference from the initial radius, $U$ is the set of tuples related to $m$. $G_{in}$ is the set of sampling line segments, which locate inside $U$ before the modification. $G_{out}$ is composed of two kinds of sampling line segments which are not inside any tuple, one is that just one terminal belongs to $U$, another is that the nearest distance to current reconstructed shape is the distance to $U$. The distance between a line segment and a tuple is defined as the bigger Hausdorff distance of two terminals to the tuple. The constraints are used to ensure that each sampling line segment is located inside at least one tuple. $E$ is the objective function that consists of a mixture of continuous and combinatorial terms: it cannot be optimized directly by gradient-based methods. Instead, we use a greedy strategy to locally optimize the configuration of each medial circle. We consider $c_o$ as the center and test positions from small radius in a search web, which is constructed by a set of concentric circles and a set of rays from their center. For each crossed node of the web, we compute the minimal $r$. Moreover, we compute the minimal value of $E$ in a defined range. Finally, we obtain the our required tuples and CTMAT, as shown in Fig. \ref{fig:CTMAT_5} and Fig. \ref{fig:CTMAT_6}. 

\begin{figure}
\subfigure[]{
\label{fig:Minkowski_1}
\includegraphics[width=0.225\columnwidth]{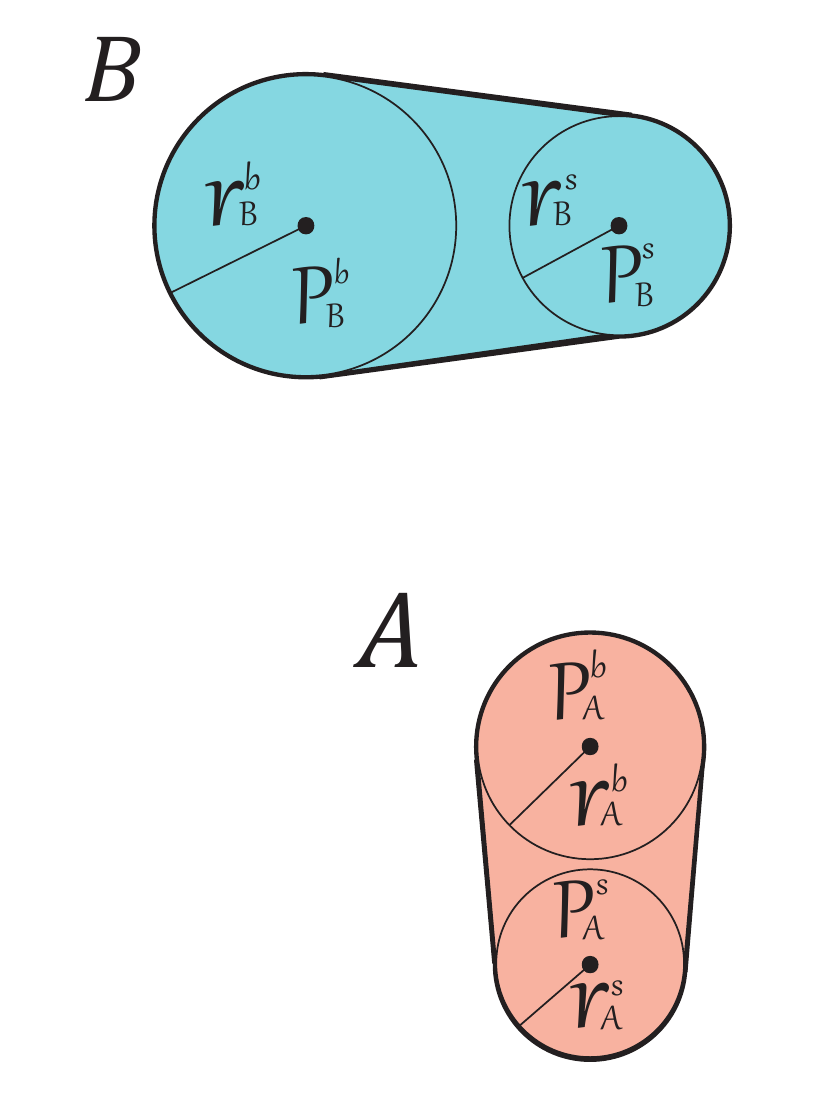}
}
\subfigure[]{
\label{fig:Minkowski_2}
\includegraphics[width=0.225\columnwidth]{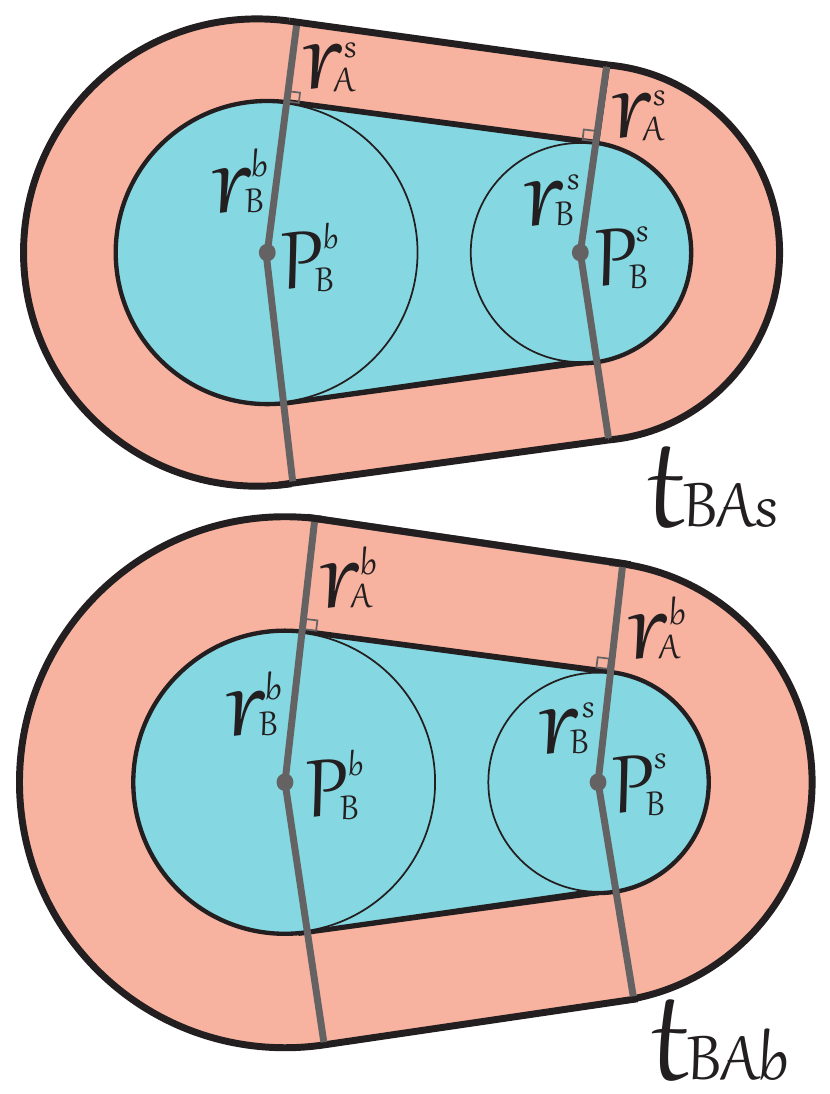}
}
\subfigure[]{
\label{fig:Minkowski_3}
\includegraphics[width=0.225\columnwidth]{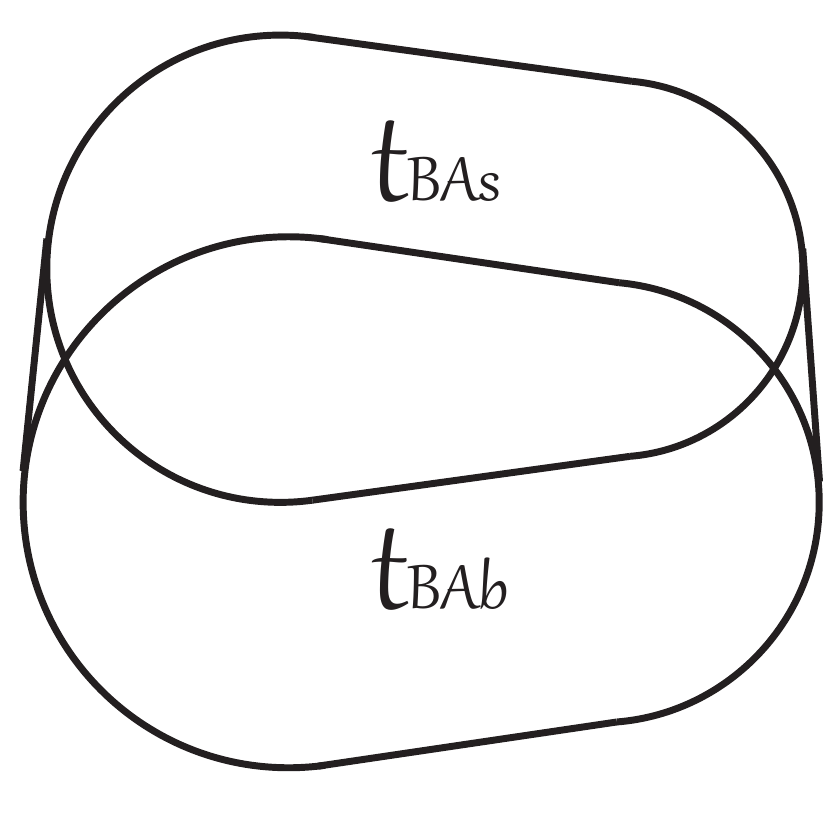}
}
\subfigure[]{
\label{fig:Minkowski_4}
\includegraphics[width=0.225\columnwidth]{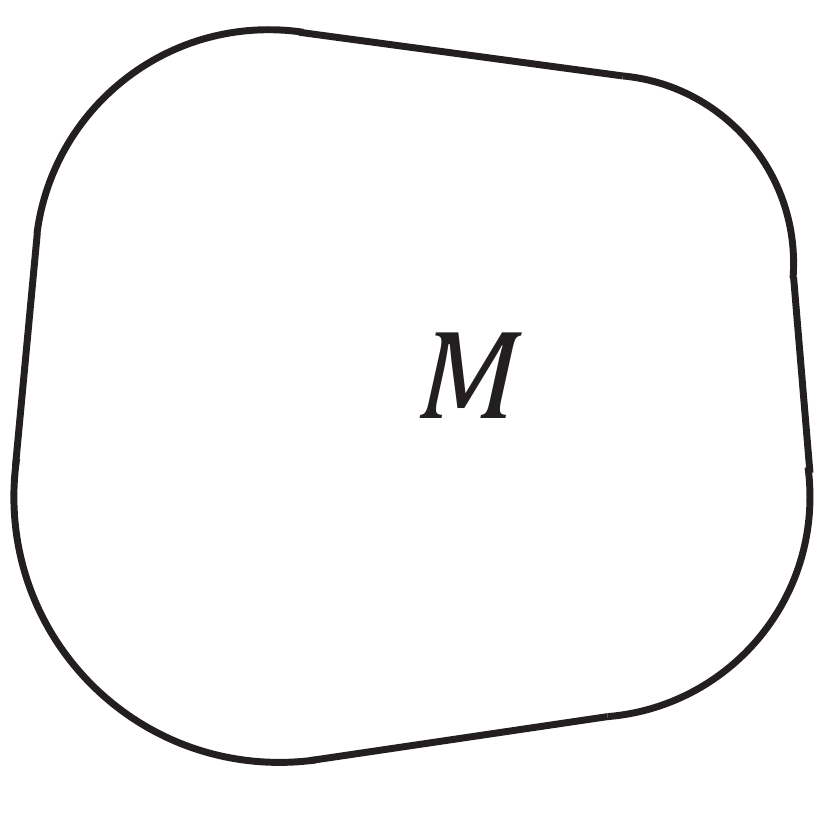}
}
\caption{Computing the Minkowski Sum of Two tuples. (a) The tuples of two agents $A$ and $B$. (b) Offsetting $B$ by two circles of $A$. (c) Positioning new tuples of (b) in right place and computing their tangent lines. (d) Getting the Minkowski Sum $M$.}
\label{fig:Minkowski}
\vspace{-2ex}
\end{figure}

\begin{figure*}
\vspace{-1ex}
\subfigure[]{
\label{fig:MARVO_1}
\includegraphics[width=0.47\columnwidth]{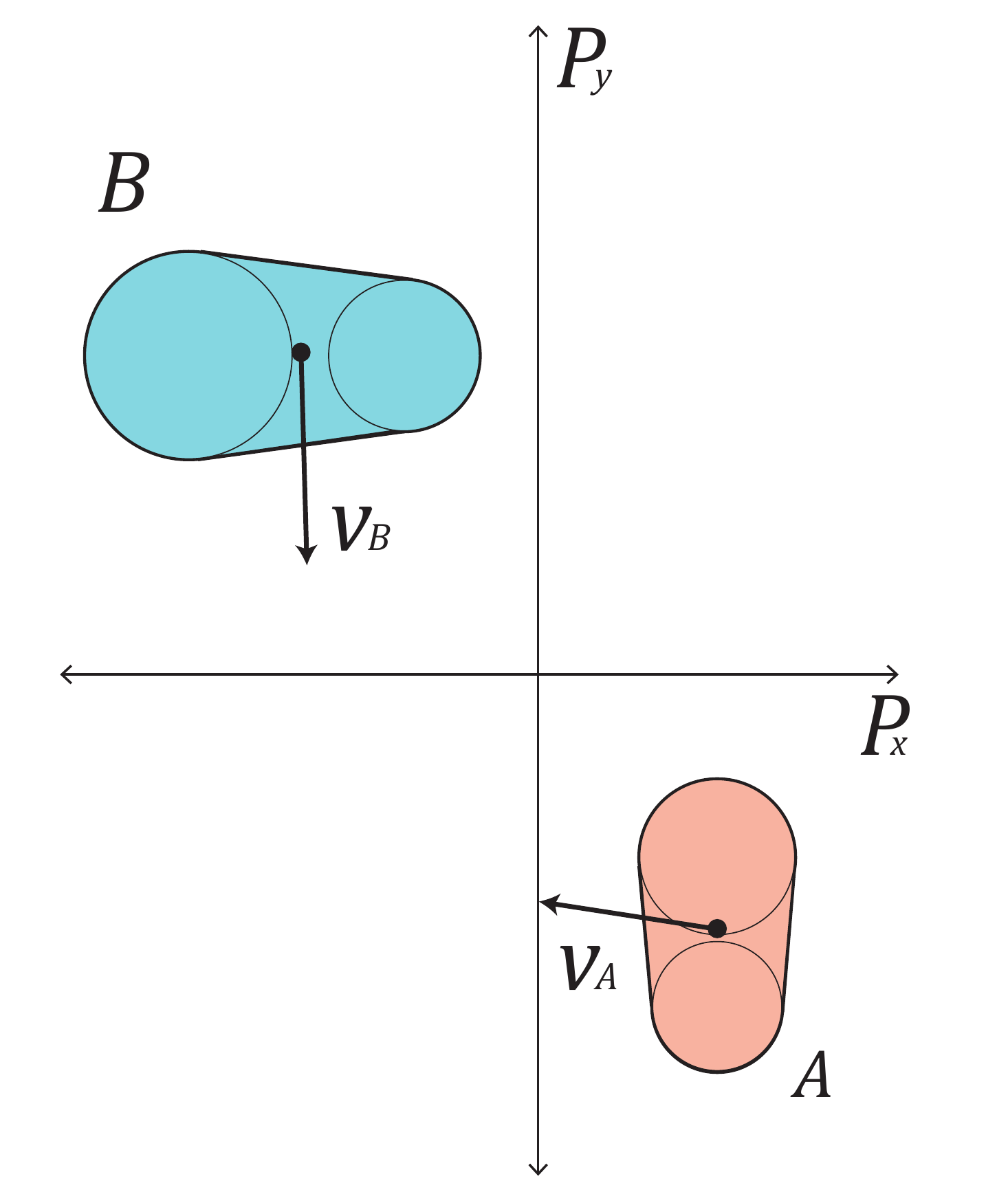}}
\subfigure[]{
\label{fig:MARVO_2}
\includegraphics[width=0.47\columnwidth]{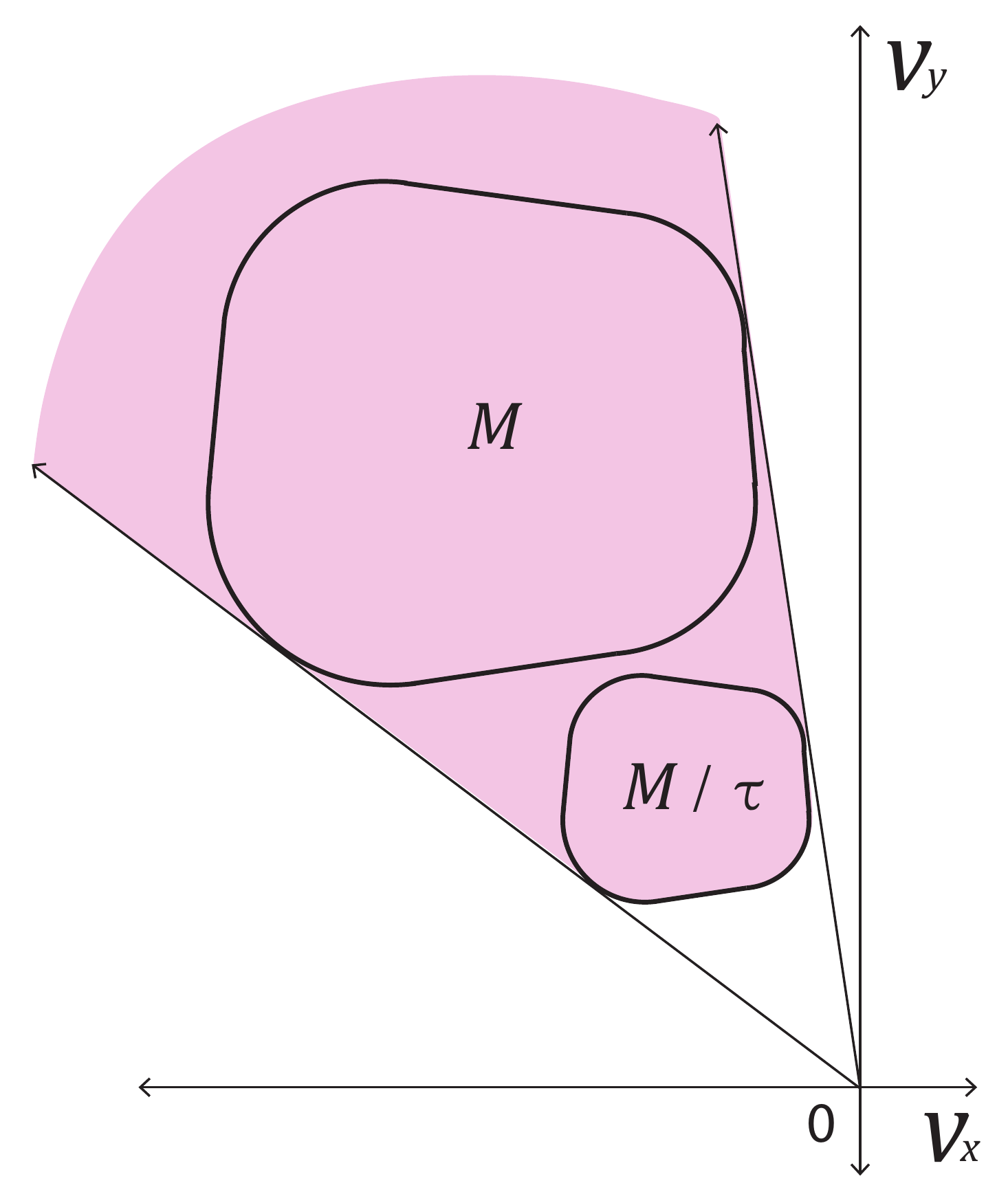}}
\subfigure[]{
\label{fig:MARVO_3}
\includegraphics[width=0.51\columnwidth]{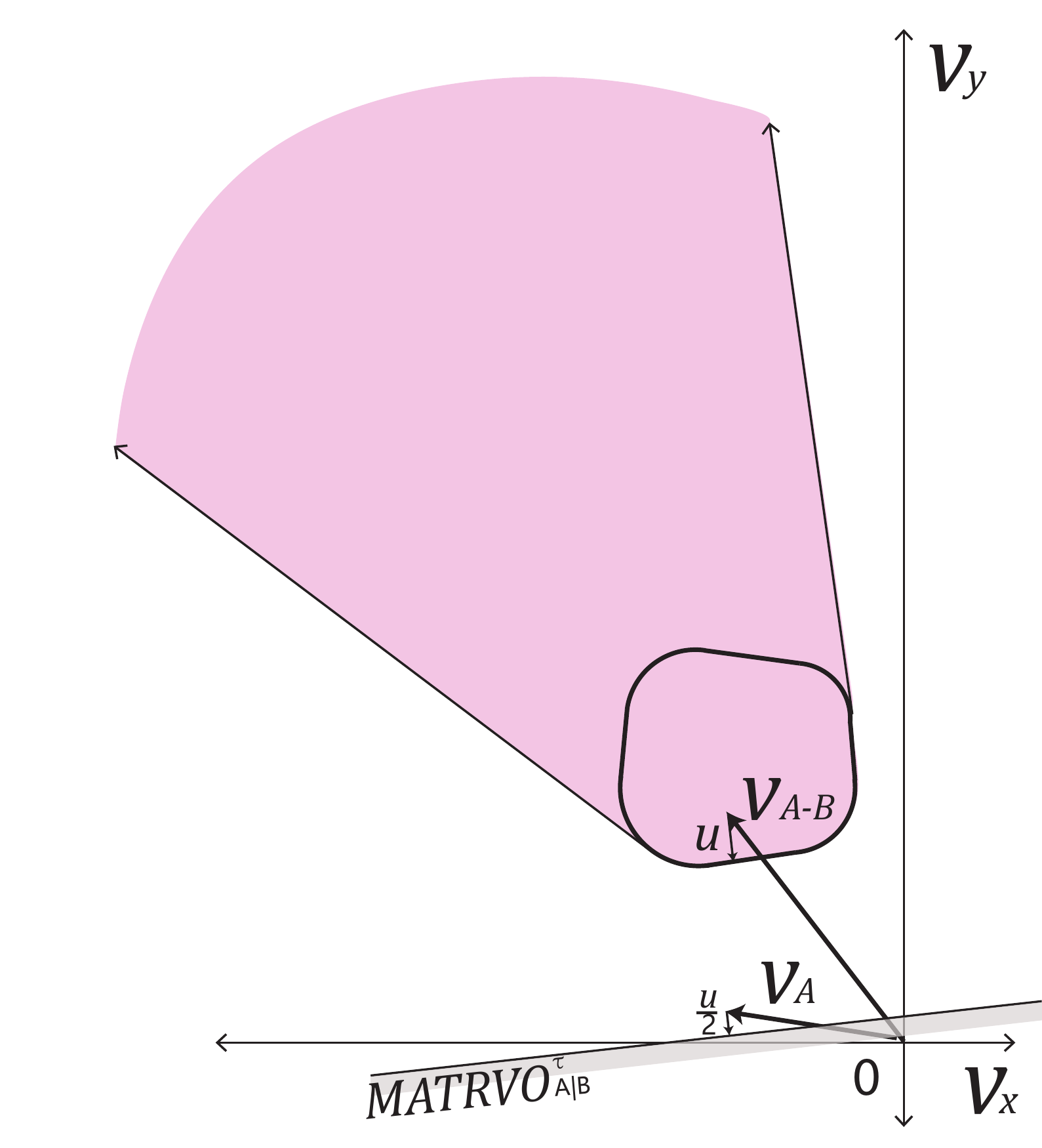}}
\subfigure[]{
\label{fig:MARVO_4}
\includegraphics[width=0.47\columnwidth]{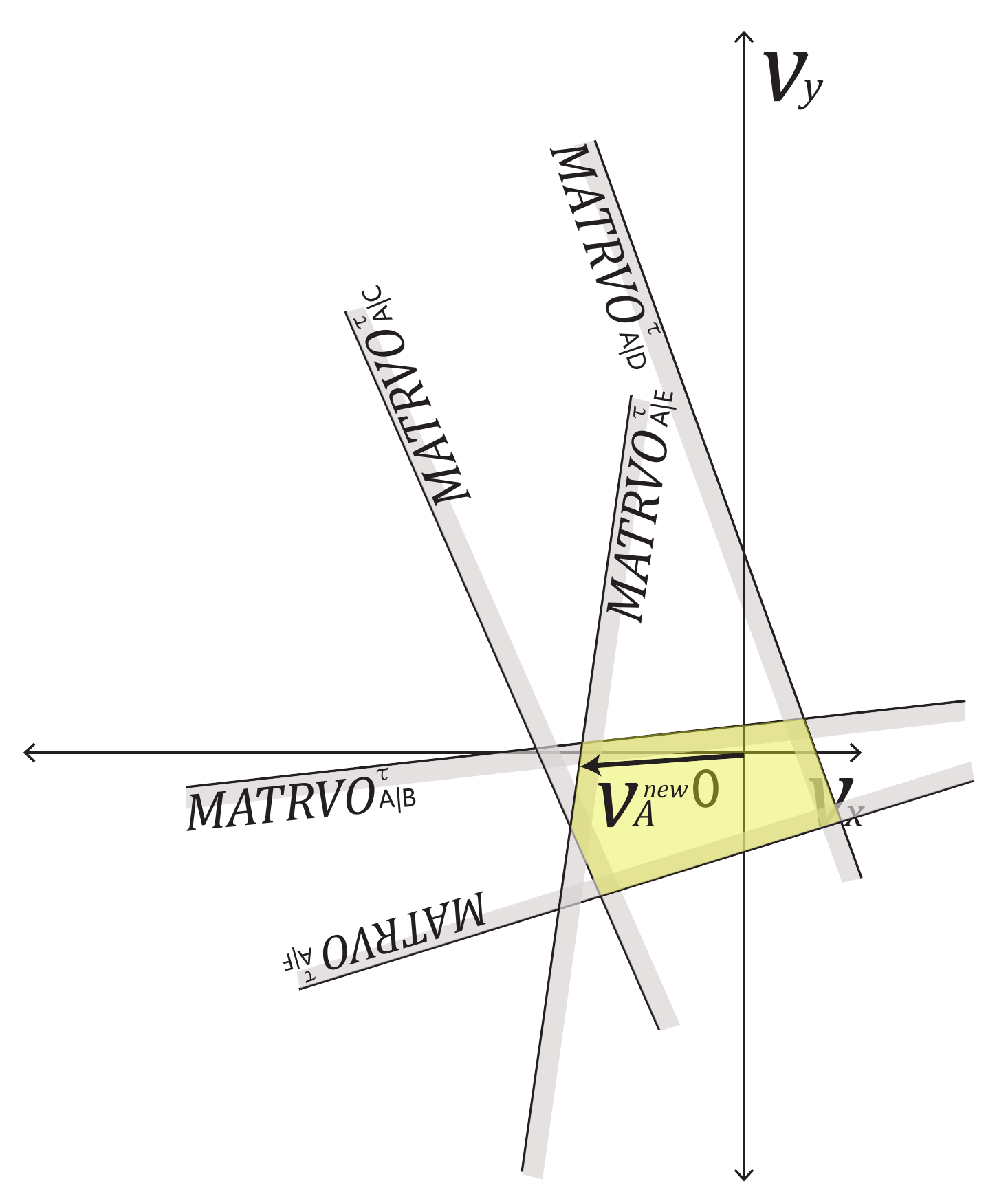}}
\vspace*{-0.1in}
\caption{MATRVO algorithm to compute collision-free velocity. (a) Two agents with one tuple are moving towards their goal positions. (b) The velocity obstacle for $A$ induced by $B$ takes the shape of a truncated cone, which is formed by the Minkowski Sum of $−A$ and $B$, scaled by $\tau$ and its two tangent lines from the origin. (c) The half-plane constraint $MATRVO_{A|B}^\tau$ is the set of permitted velocities for agent $A$ with respect to $B$. (d) After adding all the constraints, we compute a feasible velocity in the intersection region of all the half-planes.}
\label{fig:MARVO}
\vspace*{-0.1in}
\end{figure*}

\begin{theorem}
For agent $A$, $Sub(CTMAT(A)) \supseteq Sub(A)$, where $Sub$ stands for the subset of 2D space.
\end{theorem}
\begin{proof}
Let $C(A)$ represent the polygonal contour of $A$, $l$ represent the line segment on $C(A)$ and $t$ denote a tuple. For $\forall l \in C(A)$, $\exists t$, $Sub(t) \subseteq Sub(CTMAT(A))$, $l \subseteq t$. Then, $l \subseteq Sub(CTMAT(A))$. Then, $C(A) \subseteq Sub(CTMAT(A))$ and $Sub(C(A)) \subseteq Sub(CTMAT(A))$. Because $Sub(C(A)) \supseteq Sub(A)$, we get $Sub(CTMAT(A)) \supseteq Sub(A)$.
\end{proof}
Our approach can be used to represent heterogeneous agents. Fig. \ref{fig:trafficView_2} is a scene of the top view of traffic scenario with different kinds of vehicles. CTMATs for all the vehicles are shown in Fig. \ref{fig:trafficView_CTMAT}. The comparison among disc, ellipse and CTMAT in representing the part of the traffic scene inside a red rectangle is in Fig. \ref{fig:compareGeometry}, which illustrates that CTMAT provides a tighter approximation.

\section{Velocity Obstacles for Heterogeneous Agents}

In general, the physical workspace of robots or agents is in 3D, presented by $R^3$. We project the geometric representation of agents to a lower dimension $R^2$, and generate CTMAT as the underlying and conservative approximation of each agent. In order to compute collision-free velocities, we treat each tuple, which is convex, as a separate computational unit to compute the velocity of each agent. 

\subsection{Tuple Definition}
In the following, to make description clearer, we will assume the CTMAT of our agent has just one tuple. The cases that agent is represented by a few of tuples will be discussed later. The shape of tuple in $R^2$ of each agent $i$ is decided by its two medial circles - $m_b$ with bigger radius and $m_s$ with smaller radius. If two medial circles have the same radius, $m_b$ and $m_s$ are randomly assigned to them. Let position vector $\vec{p}_i^b \in R^2$ and radius $r_i^b \in R^1$ represent $m_b$ and $\vec{p}_i^s$ and $r_i^s$ represent $m_s$. In order to benefit later computation, we also store four tangent points of tuple, shown in Fig. \ref{fig:exampleTuple}, which can be represented by $T_i= \left[ T_1^i, T_2^i, T_3^i, T_4^i\right]$. The case that the computed CTMAT of agent degenerates to one circle is easier to deal with~\cite{van2011reciprocal}. We just consider common situations. So the information of the tuple of agent $i$ can be represented as $t_i = \left[ \vec{p}_i^b, r_i^b, \vec{p}_i^s, r_i^s, T_i \right]$. The preferred velocity and current velocity are denoted by $\vec{v}_i^0$ and $\vec{v}_i$ repectively. Let $\vec{o}_i^0$ denote the preferred orientation and $\vec{o}_i$ denote current orientation for the agent. Then the state space of agent $i$ is given by $\left[t_i, \vec{v}_i^0, \vec{v}_i, \vec{o}_i^0, \vec{o}_i\right]$. If the agent has more than one tuple, $t_i$ can be changed to a set of tuples.

\subsection{Local Collision Avoidance}
Our algorithm is based on Velocity Obstacle (VO)~\cite{fiorini1998motion} and its follow-up ORCA ~\cite{van2011reciprocal}. For two agents $A$ and $B$, the velocity obstacle of $A$ induced by $B$ is represented by $VO_{A|B}^\tau$, which consists of all the relative velocities of $A$ with respect to $B$ that would cause a collision with $B$ at some moment before time $\tau$. Conversely, assuming $\vec{v}_A$ and $\vec{v}_B$ are current velocity of $A$ and $B$ respectively, the condition $\vec{v}_A-\vec{v}_B \notin VO_{A|B}^\tau$ can guarantee that agent $A$ and $B$ are collision-free for at least $\tau$ time. $ VO_{A|B}^\tau$ and $ VO_{B|A}^\tau$ are symmetric in the origin. Formally,
 \begin{equation}
   VO_{A|B}^\tau = \{\vec{v} \mid  \exists t \in \left[ 0, \tau \right] :: t\vec{v} \in M \},
  \end{equation}
where $M$ is the Minkowski Sum between $B$ and $-A$. 

The tuple of $A$ and $B$ and its parameters are shown in Fig. \ref{fig:Minkowski_1}. The first step of computing $M$ is to offset $B$ by $r_A^b$ and $r_A^s$ of $A$ respectively and get two new expanded tuples $t_{BAs}$ and $t_{BAb}$, shown in Fig. \ref{fig:Minkowski_2}. The offsetting operation is composed of two substeps - one is enlarging the two circular arcs and the other is shifting two tangent lines of $B$ in terms of vector $\vec d_1 = T_1^B - P_B^b$ and $\vec d_2 = T_3^B - P_B^b$. After getting new tuples, we translate them to correct places according to the position of $-A$. $t_{BAs}$ moves according to the vector $-P_A^s$, and $t_{BAb}$ translates by the vector $-P_A^b$. The result is in Fig.\ref{fig:Minkowski_3}. We only need compute tangent lines of two new tuples and then the boundary of $M$, defined by $\Omega_M$, can be extracted. $\Omega_M$ is still composed of line segments and circular arcs, which brings convenience to later computation of nearest point and forward face. 

\subsection{Neighboring Obstacle Constraints}
It comes to compute the velocity obstacle for agent. Also taking agent $A$ and its neighboring agent $B$ as an example, we know their current velocities and their Minkowski Sum $M$. The next step is to find lines from the origin and tangent to $\Omega_M$. It is easy to compute these lines because of the geometric properties of the components of $\Omega_M$. Then we can get the tangents of $\Omega_M$ scaled by the inverse of $\tau$ and the forward face, shown in Fig. \ref{fig:MARVO_2}. The forward face is also composed of line segments or circular arcs or combination of them, so the nearest point of the boundary of velocity obstacle for the relative velocity $\vec{v}_{A-B}$ can be computed easily.

Then, we compute valid velocities for agent $A$ by making use of the velocity obstacle. The process is extended from ORCA ~\cite{van2011reciprocal} and we denote the permitted velocities for $A$ for reciprocal collision avoidance with respected to $B$ as $MATRVO_{A|B}^\tau$. As Fig. \ref{fig:MARVO_3} shows, $\vec{u}$ is the vector from $\vec{v}_{A-B}$ to the nearest point. Agent $A$ should change its velocity by $\frac{1}{2} \vec{u}$ under the assumption that $B$ will do the same. And the collision-free velocity for $A$ with the neighbor $B$ is defined by the half-plane, passing through the point $\vec{v}_A + \frac{1}{2} \vec{u}$ and vertical to $\vec{u}$. Suppose agent $A$ and its neighbor $B$ has $m$ tuples and $n$ tuples respectively, the planes caused by them is $m\times n$. If we have found all neighboring agents of $A$ by searching in kD-tree, we can compute all the half-planes, the constraints for $A$, and get the final intersected valid area $MATRVO_A^\tau$ for permitted velocities of $A$. 
\begin{equation}
  MATRVO_A^\tau = \bigcap_{B \neq A} MATRVO_{A|B}^\tau
\end{equation}
To guarantee our agent always has a tendency towards its goals, we choose a velocity inside $MATRVO_A^\tau$ that has the smallest deviation from its preferred velocity $\vec{v}_A^0$. That is
\begin{equation}
  \vec{v}_A^{new} = \argmin_{\vec{v} \in MATRVO_A^\tau}{\left\| \vec{v} - \vec{v}_A^0 \right\|  }
 \end{equation}
 We use linear programming to compute $MATRVO_A^\tau$ and Eq. 3, and the runtime is $O(n)$ where $n$ is the number of constraints. If there is a feasible solution for the agent, the agent's motion is guaranteed to be collision-free. When the agents are densely distributed in the scenario, MATRVO may be empty and no feasible solution can be found. In that situation, we minimally penetrate the constraints and use another linear programming ~\cite{van2011reciprocal, best2016real}. 
 \begin{theorem}
 If MATRVO algorithm is able to compute a feasible velocity, the resulting motion for agent is collision-free.
\end{theorem}
\begin{proof}
For $\forall$agent $A$ in the scenario, its neighboring agents can be represented as $Neighbors(A)$. All the tuples of $A$ construct constraints with all the tuples of $Neighbors(A)$. If we compute a feasible velocity $\vec{v}_A^{new} \in MATRVO_A^\tau$ by linear programming, then $\forall$tuple $t_n \in Neighbors(A)$, $\forall$tuple $t_A \in A$, $\vec{v}_A^{new} \notin VO_{t_A|t_n}^\tau$. Then, $\forall$agent $N \in Neighbors(A)$, $\vec{v}_A^{new} \notin VO_{A|N}^\tau$. Then, $A$ will not collide with any other neighboring agent at any moment before time $\tau$. 
\end{proof}

\subsection{Runtime Analysis}
Let $t(A)$ represent the number of tuples of $CTMAT(A)$ and $N_i^A$ represent the $i$th neighboring agent of $A$. The runtime $Time(A)$ of computing $MATRVO_A^\tau$ can be denoted as follows.
\begin{equation}
  Time(A) = t(A) \times \sum_{k=1}^nt(N_k^A) \times \mu,
 \end{equation}
where $n$ denotes the number of neighboring agents of $A$, $\mu$ is the runtime of computing $MATRVO_A^\tau$ when $A$ has just one neighbor and both of them have one tuple.

\begin{figure}
\vspace{-2ex}
\subfigure[]{
\label{fig:precomputation_2}
\includegraphics[width=0.45\columnwidth]{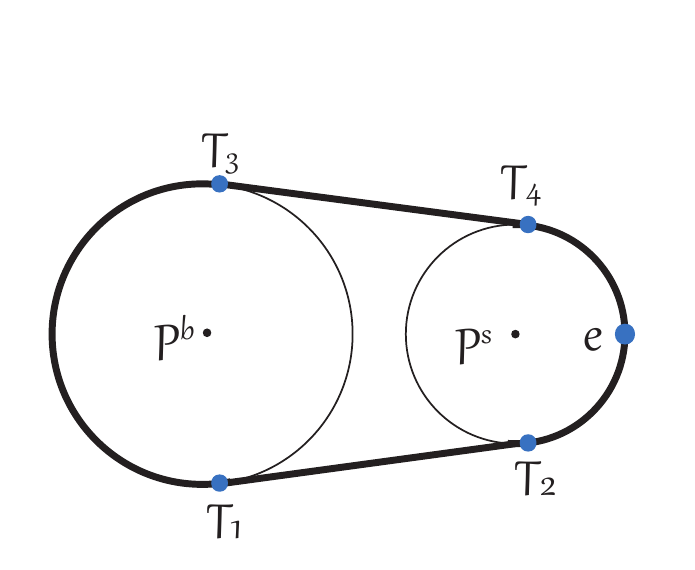}}
\subfigure[]{
\label{fig:precomputation_1}
\includegraphics[width=0.45\columnwidth]{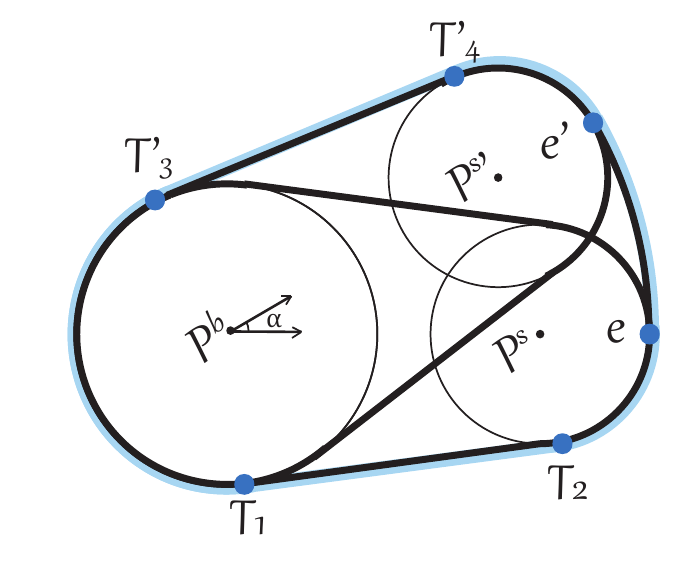}}
\vspace*{-0.1in}
\caption{Swept Tuple. (a) An original tuple. (b) Rotate the tuple by an angle $\theta$ with $P^b$ as the pivot. $e$ is the middle point on circular arc $T_2T_4$ of the original tuple and the corresponding point after rotating is represented as $e^\prime$.}
\label{fig:precomputation}
\vspace*{-2ex}
\end{figure}

\begin{figure}
\vspace{-2ex}
\subfigure[]{
\label{fig:orientation_1}
\includegraphics[width=0.42\columnwidth]{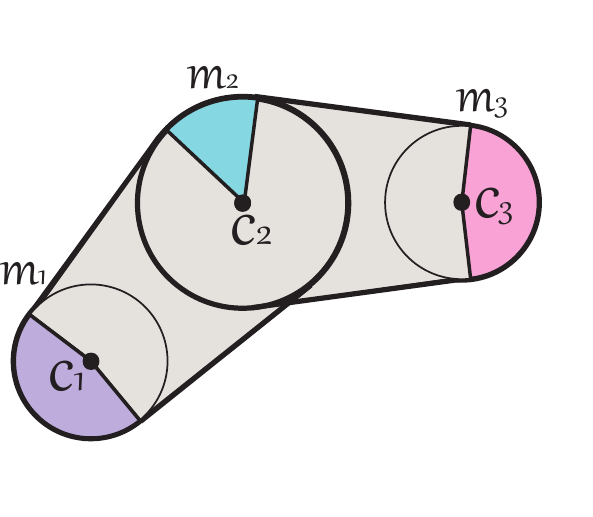}}
\subfigure[]{
\label{fig:orientation_2}
\includegraphics[width=0.42\columnwidth]{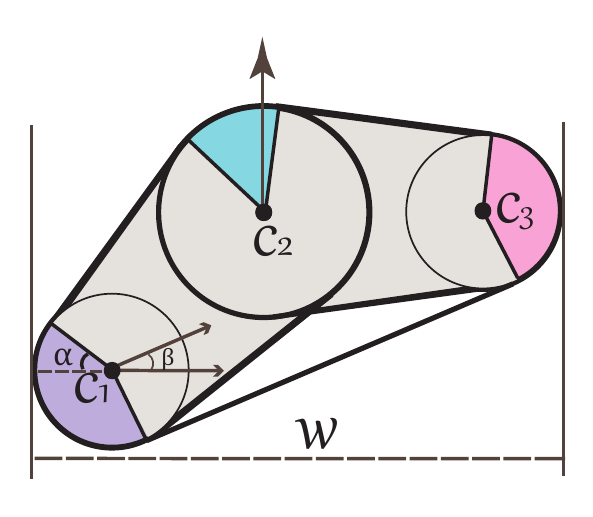}}
\vspace*{-0.1in}
\caption{Computing the width for agent: (a) An agent with two tuples. (b) Convex hull of the agent. $w$ is the width of the orientation of the arrow.}
\label{fig:orientation}
\vspace*{-2ex}
\end{figure}

\begin{figure*}
\vspace{-1ex}
\subfigure[]{
\label{fig:t_view1}
\begin{minipage}[b]{0.185\textwidth}
\includegraphics[width=1\textwidth]{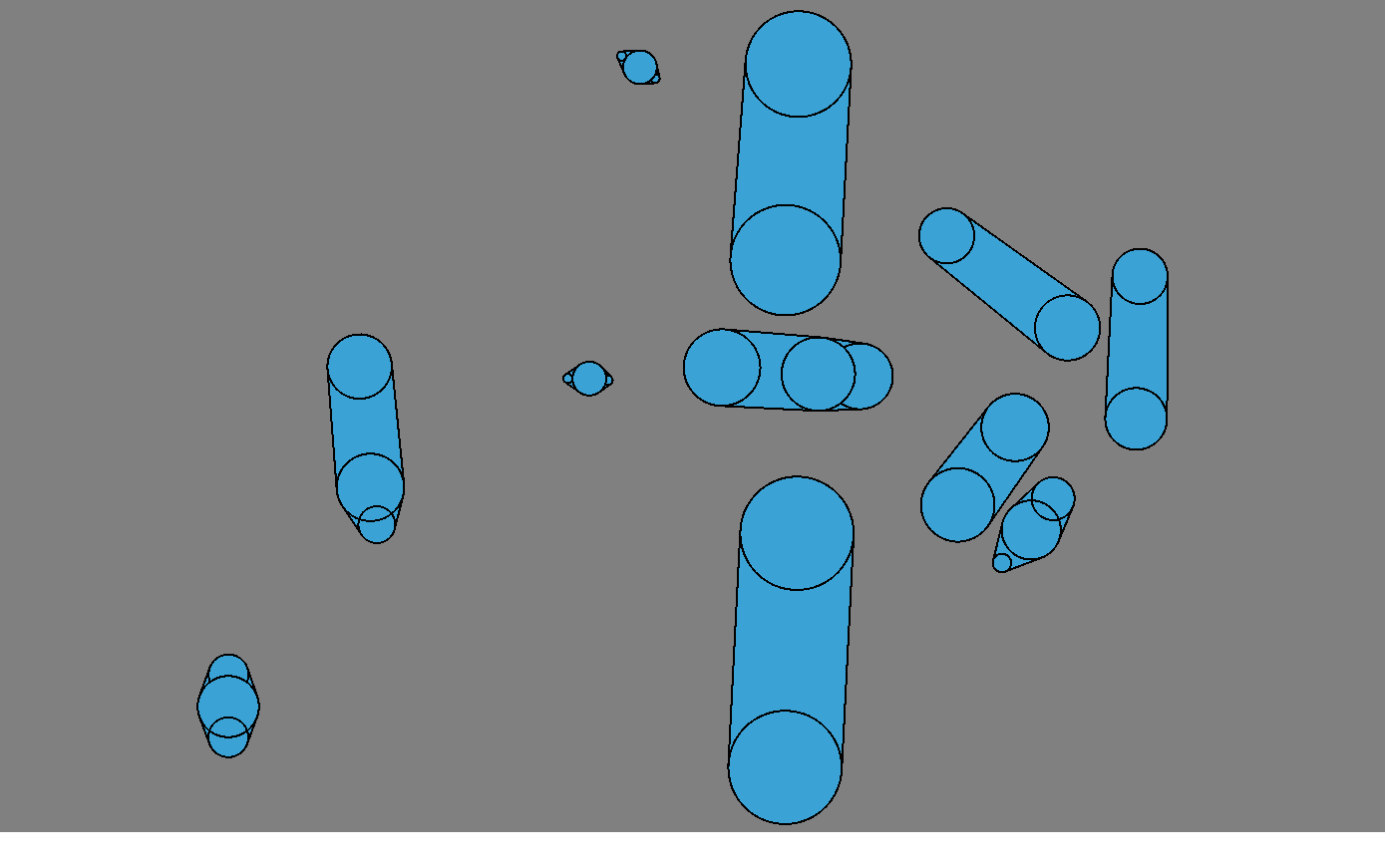} \\
\includegraphics[width=1\textwidth]{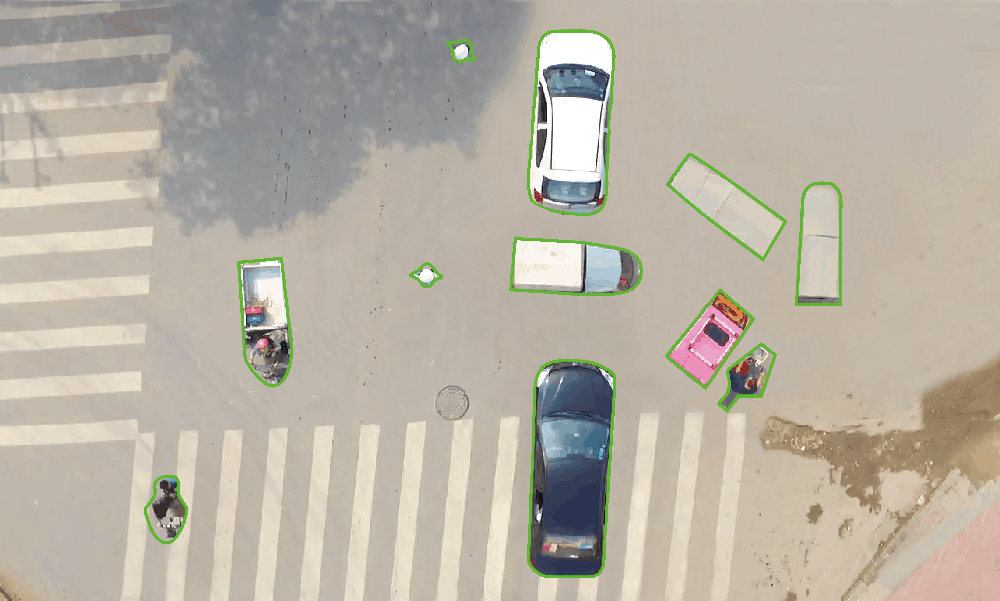}
\end{minipage}}
\subfigure[]{
\label{fig:t_view2}
\begin{minipage}[b]{0.185\textwidth}
\includegraphics[width=1\textwidth]{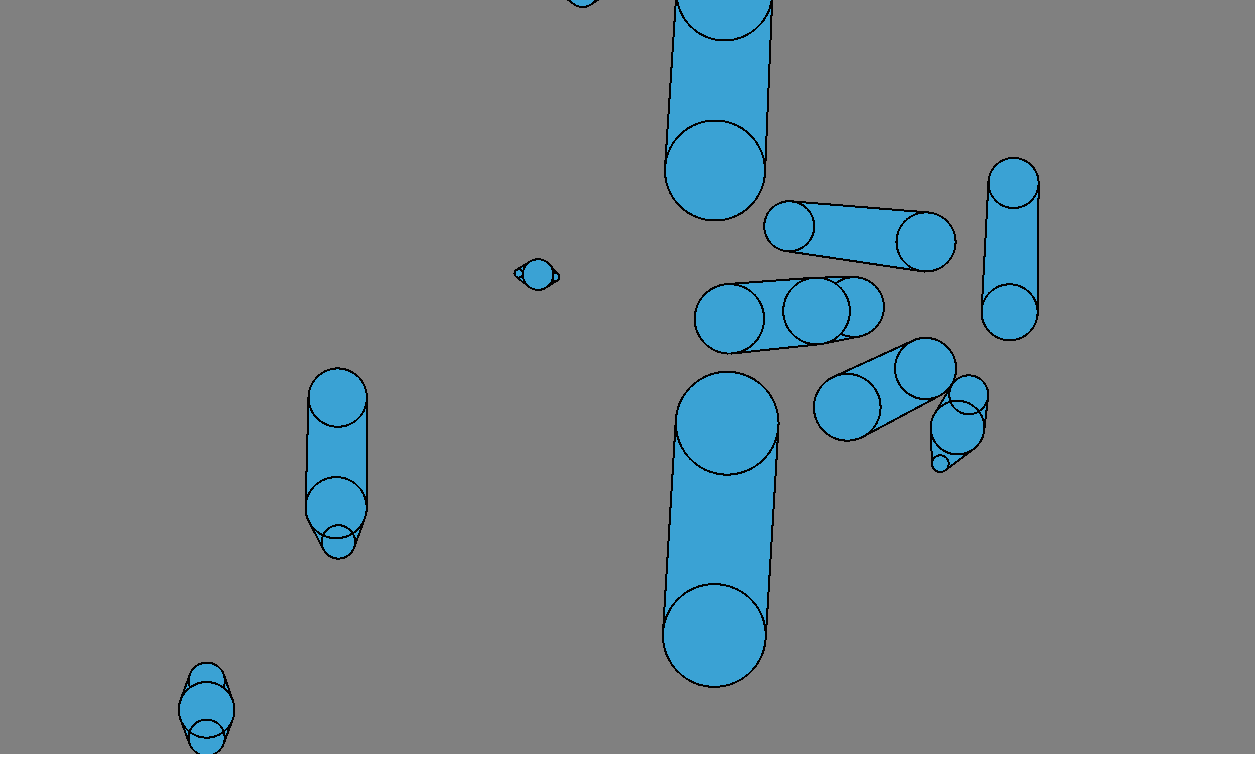} \\
\includegraphics[width=1\textwidth]{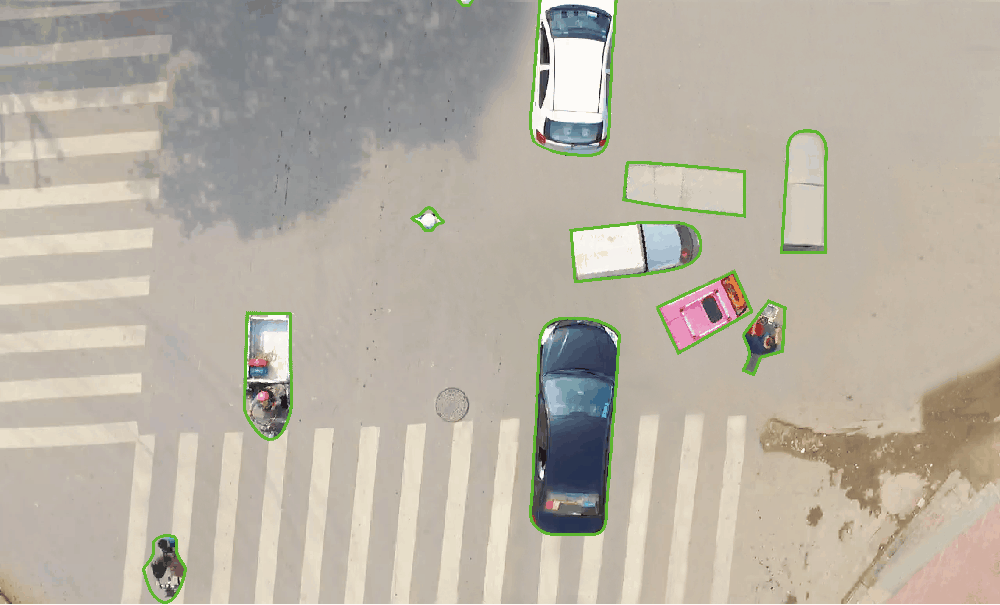}
\end{minipage}}
\subfigure[]{
\label{fig:t_view3}
\begin{minipage}[b]{0.185\textwidth}
\includegraphics[width=1\textwidth]{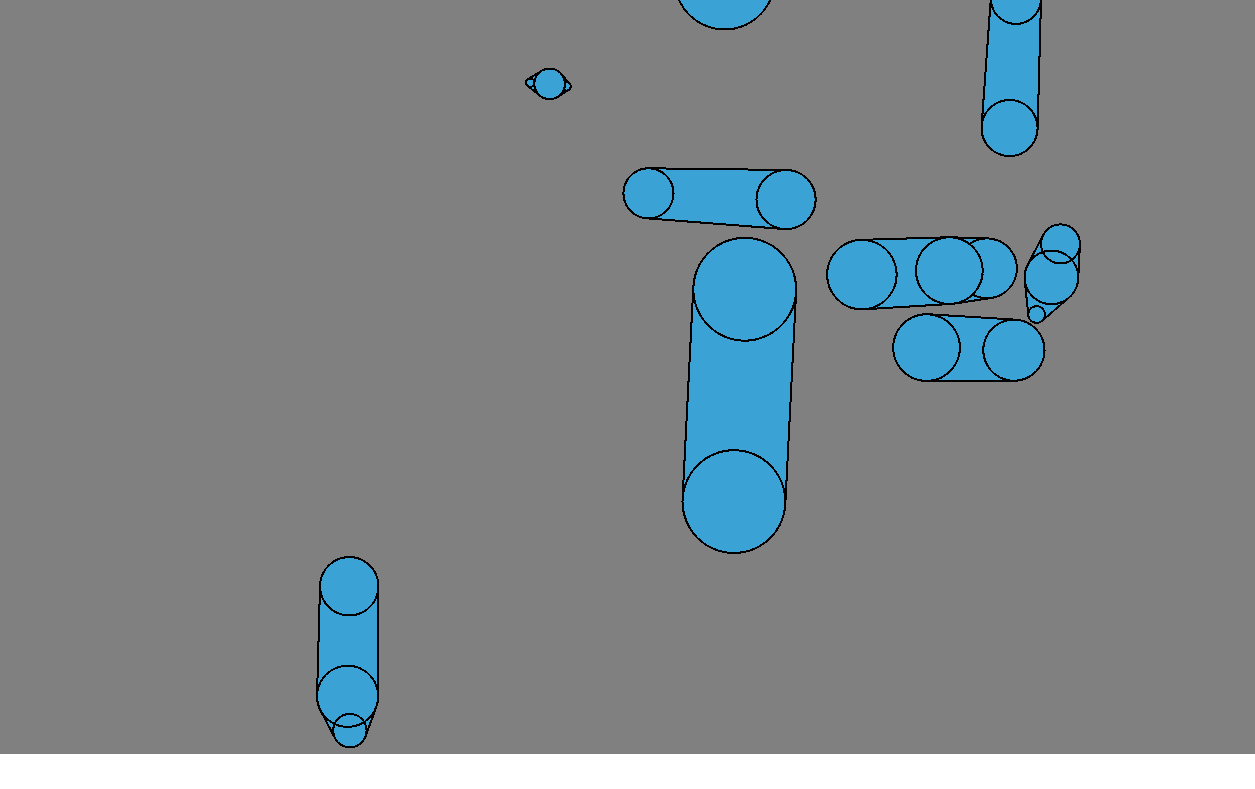} \\
\includegraphics[width=1\textwidth]{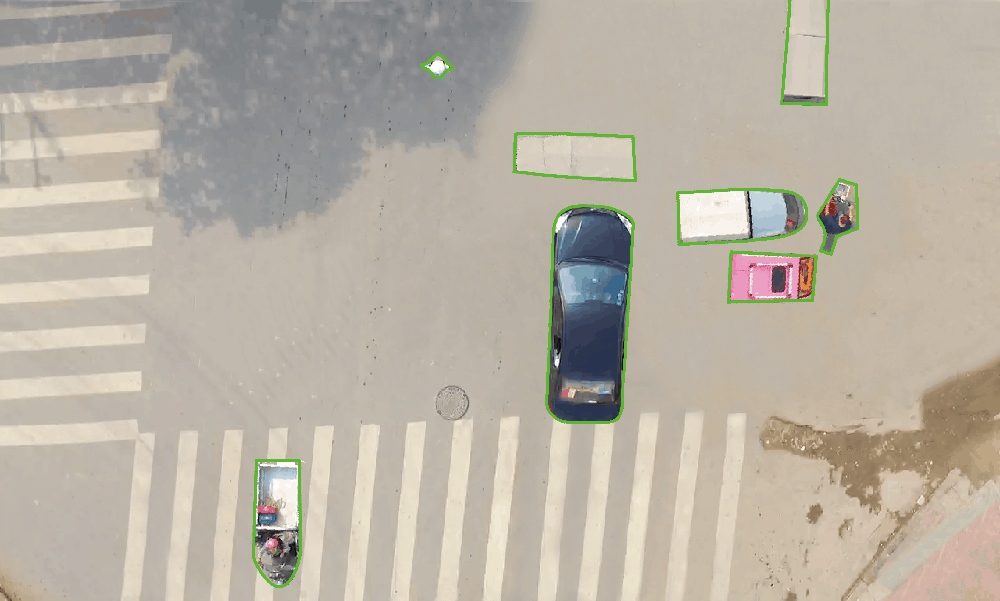}
\end{minipage}}
\subfigure[]{
\label{fig:t_view4}
\begin{minipage}[b]{0.185\textwidth}
\includegraphics[width=1\textwidth]{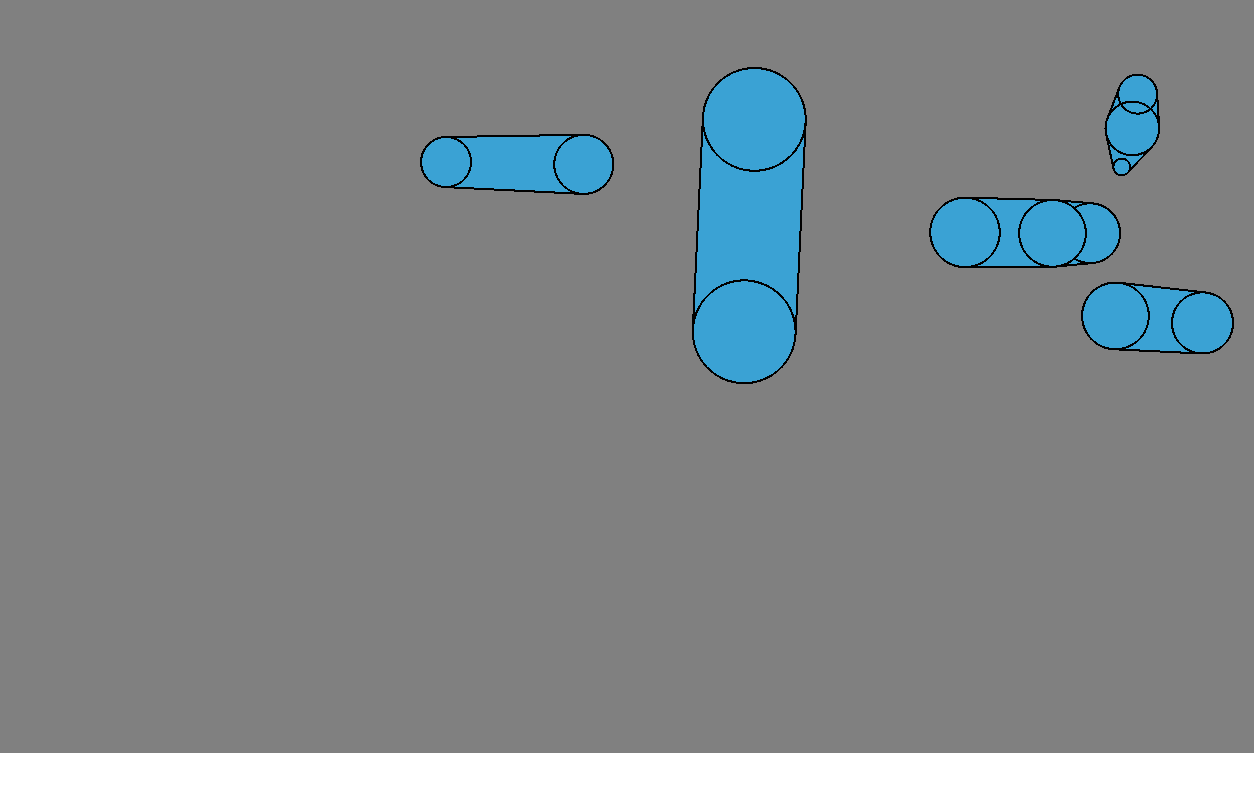} \\
\includegraphics[width=1\textwidth]{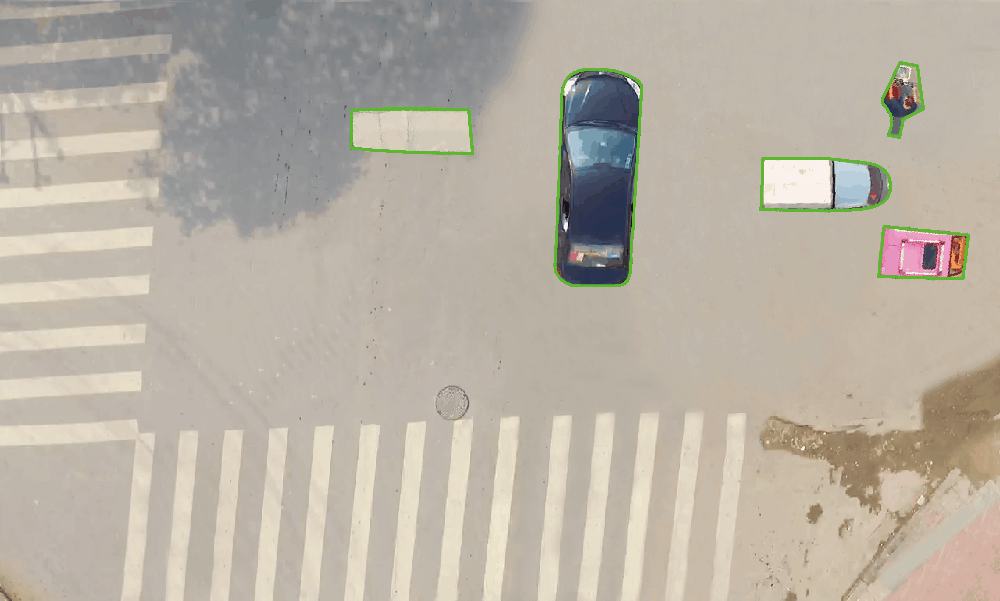}
\end{minipage}}
\subfigure[]{
\label{fig:t_view5}
\begin{minipage}[b]{0.185\textwidth}
\includegraphics[width=1\textwidth]{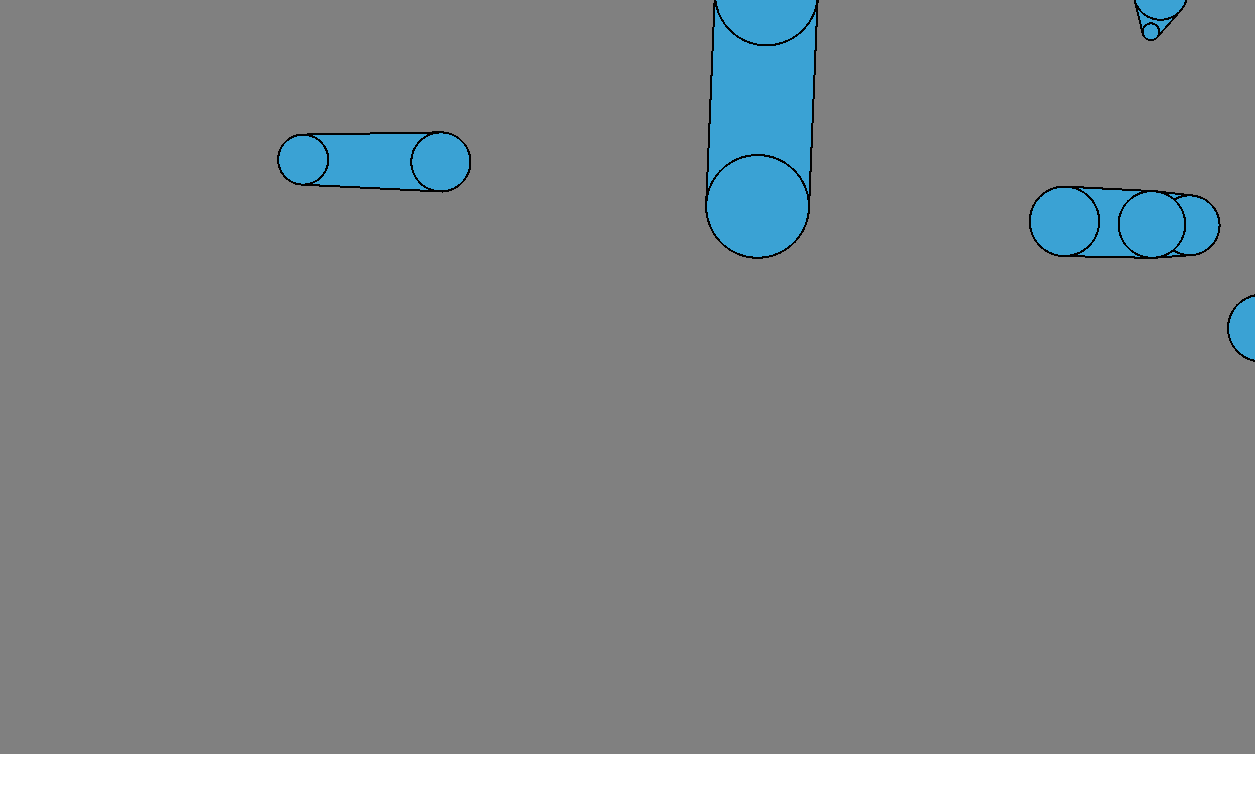} \\
\includegraphics[width=1\textwidth]{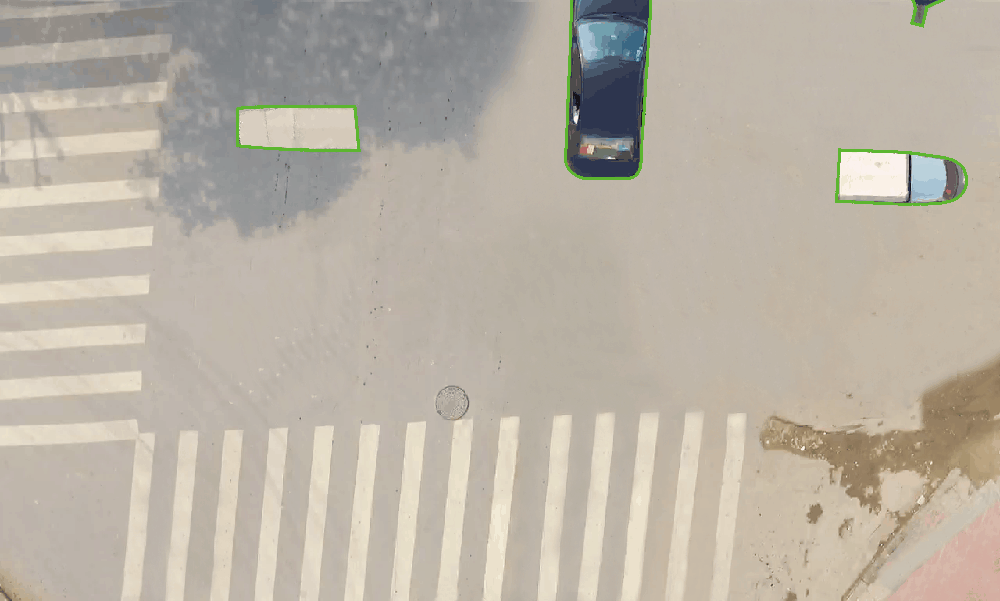}
\end{minipage}}
\vspace*{-2ex}
\caption{A sequence of frames in the simulation of traffic scenario by MATRVO.  For each scenario we compute the new velocity of each agent using MATRVO and compute collision-free trajectories between two frames.}
\label{fig:trafficview}
\vspace*{-2ex}
\end{figure*}

\begin{figure*}
\subfigure[]{
\label{fig:circle1}
\includegraphics[width=0.154\textwidth]{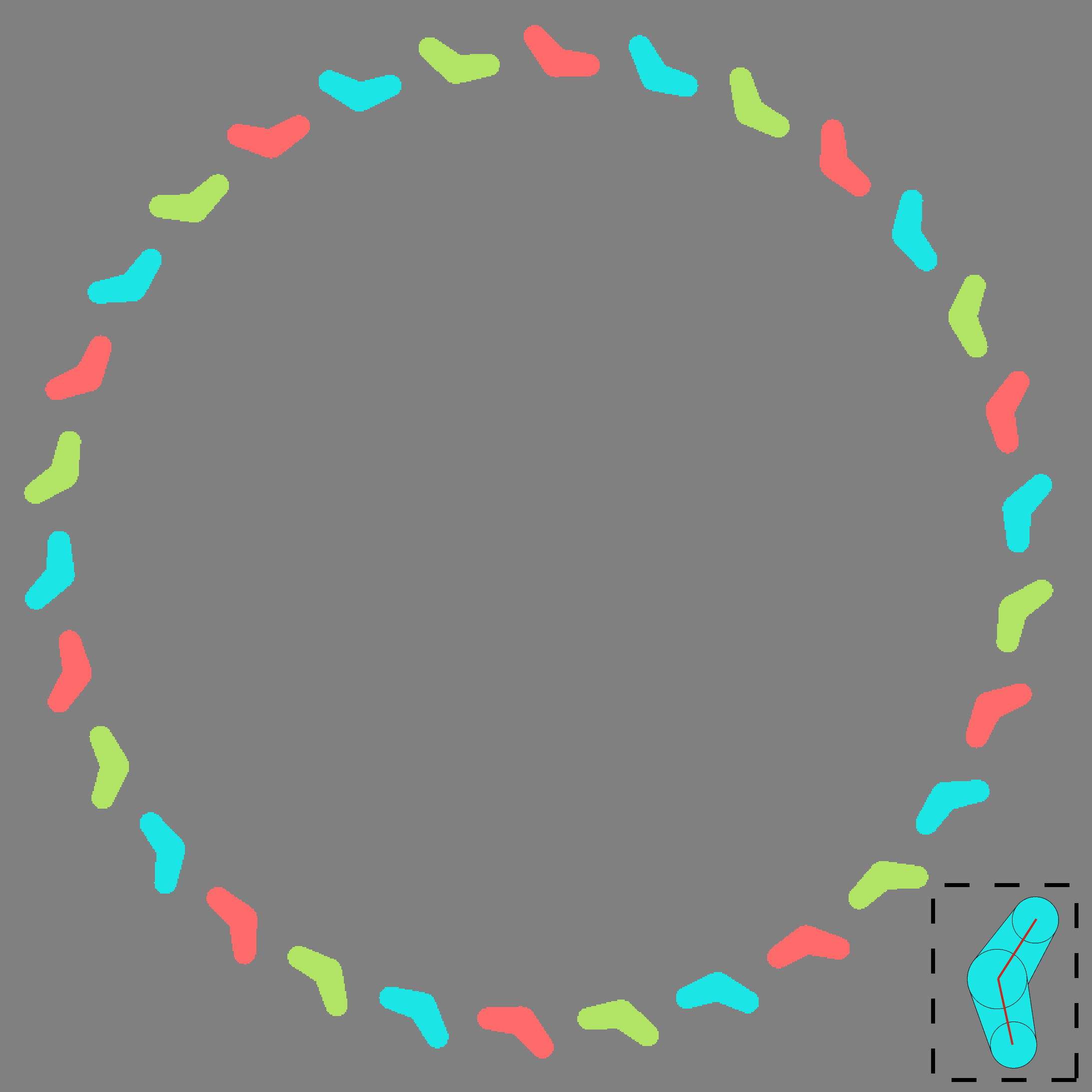}}
\subfigure[]{
\label{fig:circle2}
\includegraphics[width=0.154\textwidth]{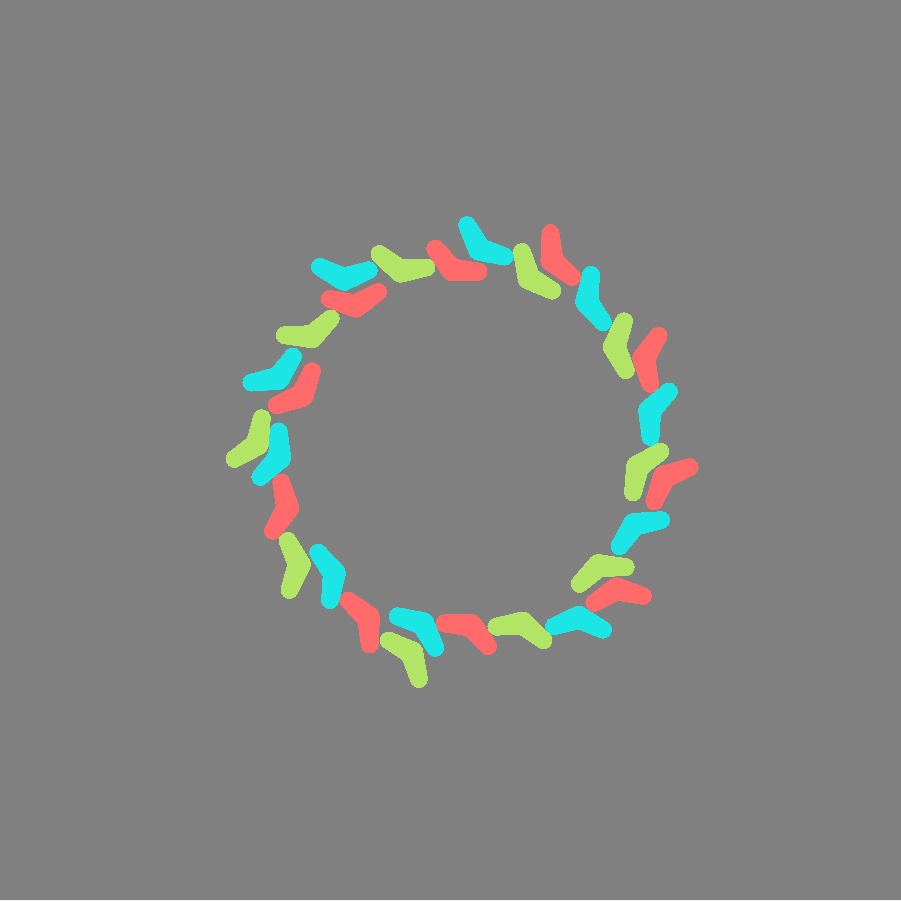}}
\subfigure[]{
\label{fig:circle3}
\includegraphics[width=0.154\textwidth]{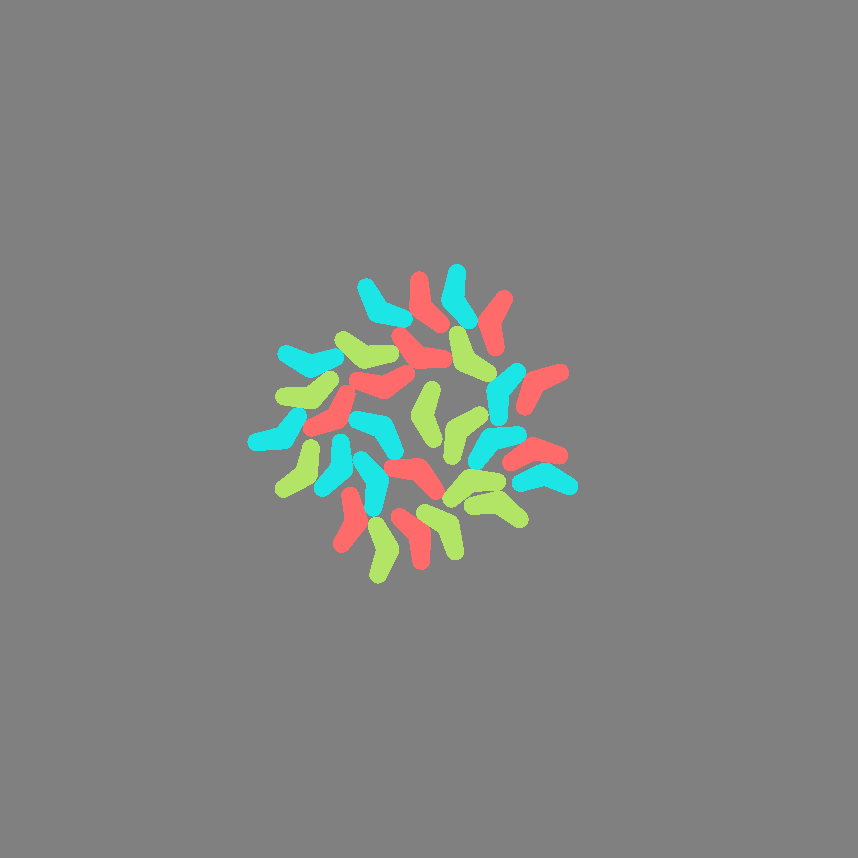}}
\subfigure[]{
\label{fig:circle4}
\includegraphics[width=0.154\textwidth]{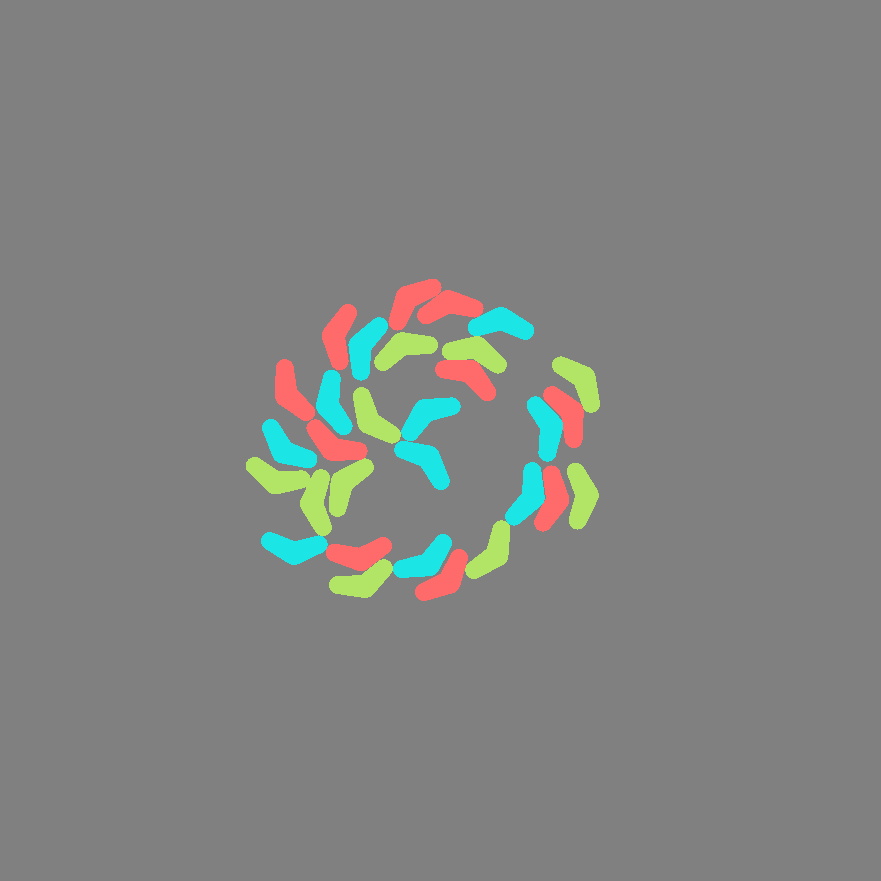}}
\subfigure[]{
\label{fig:circle5}
\includegraphics[width=0.154\textwidth]{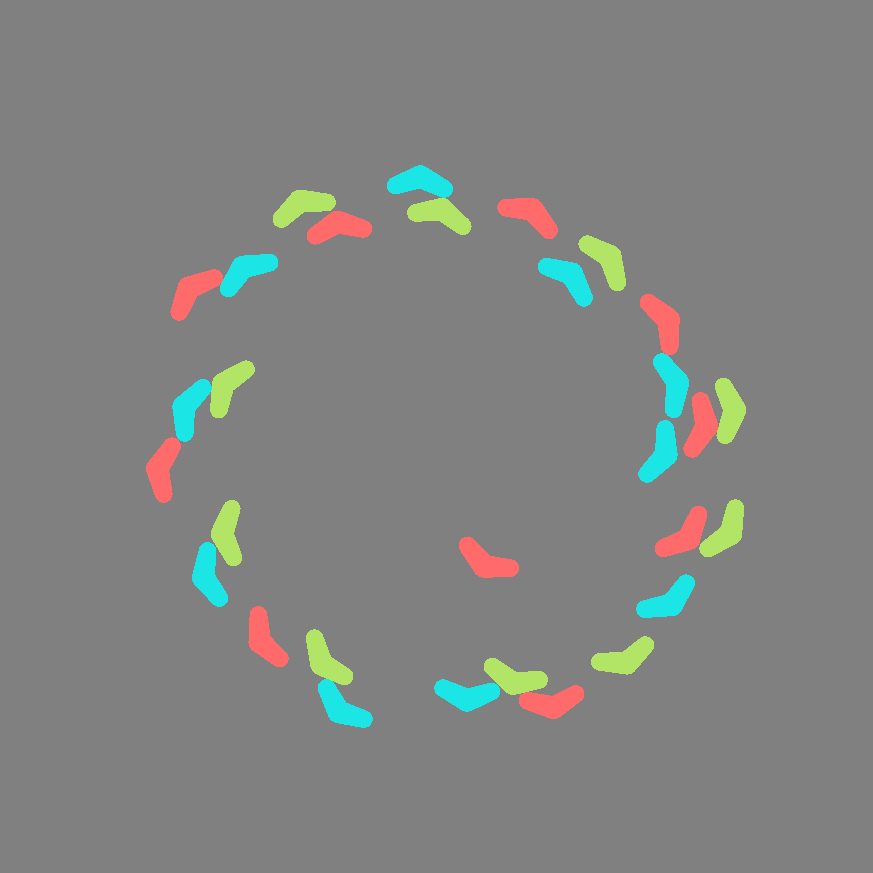}}
\subfigure[]{
\label{fig:circle6}
\includegraphics[width=0.154\textwidth]{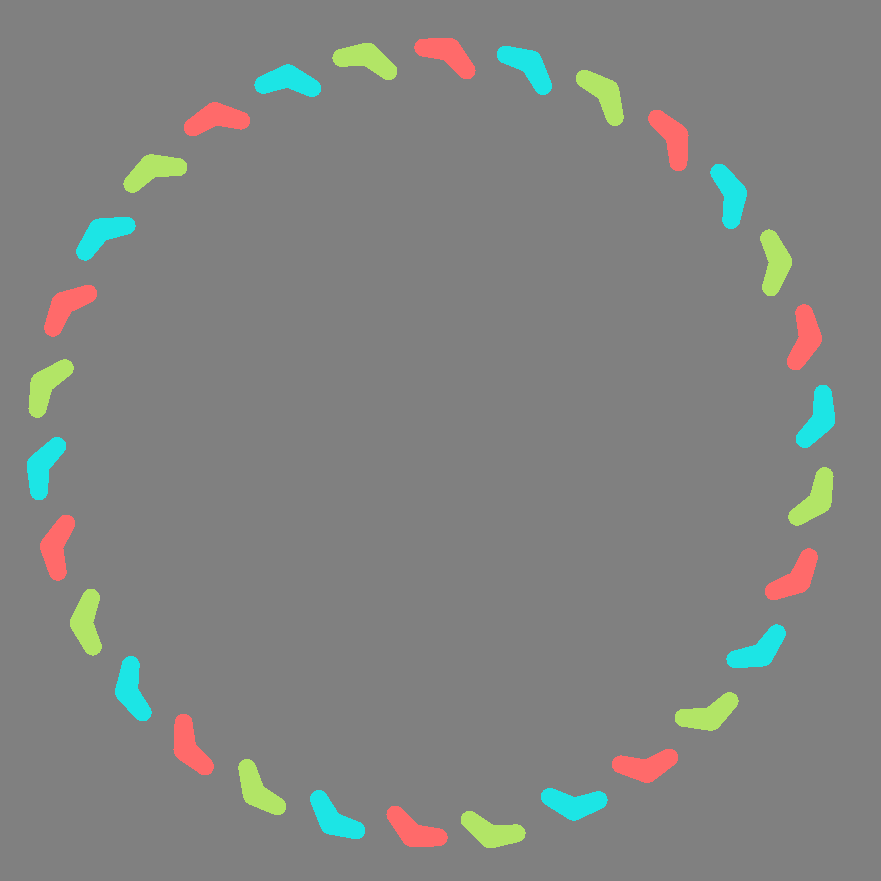}}
\vspace*{-2ex}
\caption{A sequence of frames in the simulation of antipodal circle scenario by MATRVO.}
\label{fig:antipodal}
\vspace*{-2ex}
\end{figure*}

\section{Precomputation of Minkowski Sums}
One important factor influencing the efficiency is the computation of the Minkowski Sum. To further accelerate our algorithm, we make use of a precomputational table of Minkowski Sums~\cite{best2016real}.

Given a tuple, we define the orientation angle $\theta_{t_i}$, which is the angle between the x-axis and its main axis decided by two centers of circular arcs. To make our discrete table cover all the orientations of tuples, we first compute a new representation of the tuple after rotating it for an interval angle $\alpha$. Like Fig. \ref{fig:precomputation} shows, we select $P^b$ as the rotation center because the approximation error for the tuple is smaller than using $P^s$ when their radii are different. The blue contour in Fig. \ref{fig:precomputation_1} is composed of new circular arcs and line segments. The special one is the circular arc $ee^\prime$, which does not come from the original medial circles but from a new circle with $P^b$ as the center and $\left\| P^b P^s \right\| + r^s $ as the radius. However, it does not effect our computation of the Minkowski Sum since it is also a circular arc. The structure has no essential difference with the tuple. Let $S = \left\{ {\theta}_E \times i : 0 \leq i \leq \lfloor \frac{2\pi}{\theta_E} \rfloor \mid \theta_E \in (0, 2\pi) \right\}$ denote the set of angles and $R  = \left\{ Rot(T(\theta_i), \theta_{i+1}) \mid \theta_i, \theta_{i+1} \in S \right\}$ represent the set of precomputed contours by rotating a tuple from $\theta_i$ to $\theta_{i+1}$ for each two ordered angles in $S$. After getting the set $R$ for each kind of tuple, we could construct $n\times n$ tables of Minkowski Sum for $n$ tuples. When there comes a tuple with orientation $\theta_{now}$, we can easily get the corresponding element in $R$ by searching $\theta_i \leq \theta_{now} < \theta_{i+1}$. In this way, for a pair of tuples, we can efficiently find the corresponding Minkowski Sum in the precomputed tables. In our experiment, we set $\theta_E$ to $\frac{1}{36}\pi$.

After computing the Minkowski sum using the table, we use our method to compute the forward face and nearest point. Our precomputed method provides $2\times$ improvement in runtime performance, when each agent is represented using one tuple.

\section{Orientation Update}

Our representation of agents provides more degree of freedom in terms of selecting a feasible motion to avoid collisions. When the space in the environment is too narrow for the agent to move ahead, the agent can rotate its body to find a relatively small width under the direction of its velocity to pass through the space. In order to perform these computations efficiently, we use a precomputed width table to search for the rotated angle at runtime.

In order to design a solution for general scenarios, we assume the agent has two tuples, as shown in Fig. \ref{fig:orientation_1}. The first step is to add more tangent lines to obtain the convex hulls of the agent, which consists of circular arcs and line segments. The width of the convex hull corresponds to the width of agent. The arrow in Fig. \ref{fig:orientation_2} shows the orientation of the agent and the distance $w$ between two parallel tangent lines of the convex hull in the direction of the arrow is the width of current agent. Three medial circles in this example are $m_1$, $m_2$ and $m_3$ with centers $c_1$, $c_2$ and $c_3$, respectively. Assuming two parallel tangent lines are tangent to medial circle $m_i$ and $m_j$, we can compute the value of $w$ by $ w = r_i+r_j+d(c_i, c_j)\times \cos\beta$, where $r_i$ and $r_j$ are the radii of $m_i$ and $m_j$, respectively, $d(c_i, c_j)$ is the distance between two circles' centers and angle $\beta$ is the acute angle between the vertical direction of orientation and the line passing through $c_i$ and $c_j$. This equation corresponds to the case when two parallel tangent lines are tangent to the same medial circle. If the orientation arrow rotates by $360^{\circ}$, the change of width can be represented as a piecewise continuous function with clear ranges. We precompute the function in a table so that our algorithm can search for the width and find the minimal rotated angle for the agent to pass through the clearance efficiently. We update the orientation after every time step to compute the new velocity for the agent and use the approach in~\cite{narang2017interactive} to guarantee that the rotating action is collision-free. More details are given in the report~\cite{2017ctmat}.

\section{Results}
In this section, we highlight the performance of our algorithm on different benchmarks and compare its performance with prior multi-agent navigation algorithms. We implemented the algorithm and conducted experiments in C++ on a Windows 10 laptop with Intel i7-6700 CPU and 8GB RAM. Our algorithm can be easily parallelized on multiple cores. All the results in this paper were generated on a single CPU core.

Fig. \ref{fig:trafficview} shows a sequence of agent positions (corresponding to different vehicles) of the simulation result of a traffic scenario by MATRVO. We have computed a representation and position for each vehicle based on the discrete time instances in a given video. For each column, the top image corresponds to the scene of moving vehicles and the bottom one is the corresponding simulated traffic scene. Our algorithm can tightly represent different kinds of agents (vehicles) and closely match the actual traffic. Fig. \ref{fig:antipodal} shows a sequence of frames of simulating the agent positions in the antipodal scenario that has been used in prior benchmarks~\cite{van2011reciprocal,best2016real}. This scenario requires every agent on a circle to reach the antipodal position as the final goal. The type of agent used in this benchmark is shown in the dotted rectangle. We also use another four scenarios to test the performance of our algorithm. The result in Fig. \ref{fig:result_2} illustrates that our algorithm could be applied to large scenarios with hundreds of or thousands of agents and used for interactive navigation. In Fig. \ref{fig:result_3}, two agents should pass through the narrow hallway and reach the opposite positions. In this example, no disc-based agent or ellipse-based agent could pass through it due to the narrow space. Our CTMAT representation works well in such scenarios because of a tighter and more flexible representation. Fig. \ref{fig:result_4} and Fig. \ref{fig:result_5} show the performance of multi-agent navigation among static obstacles.

\begin{figure}
\vspace{-1ex}
\includegraphics[width=0.85\columnwidth]{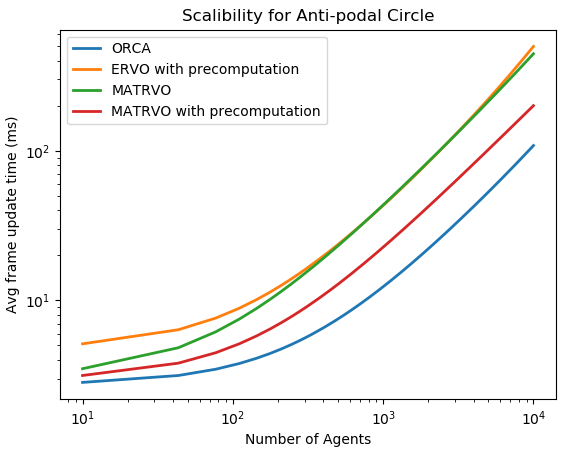}
\caption{Performance comparison of ORCA, ERVO with precomputation, MATRVO and  MATRVO with precomputation in the antipodal circle scenario.}
\label{fig:comparisonOne}
\vspace*{-1ex}
\end{figure}

\begin{figure}
\vspace{-1ex}
\includegraphics[width=0.85\columnwidth]{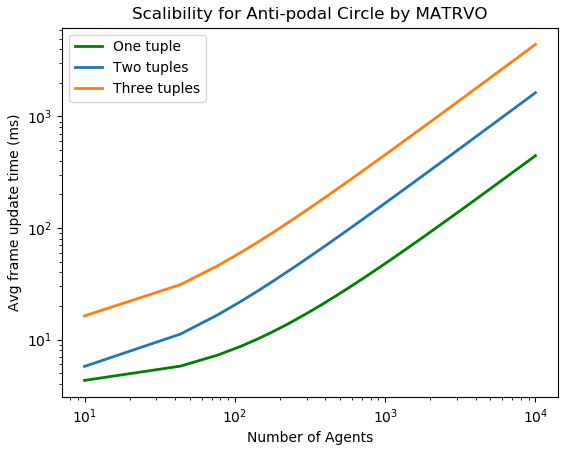}
\caption{Performance comparison of MATRVO without precomputation, when underlying agents have different numbers of tuples in the CTMAT representation.}
\label{fig:comparisonTwo}
\vspace*{-1ex}
\end{figure}

\begin{figure}
\subfigure[]{
\label{fig:result_2}
\includegraphics[width=0.47\columnwidth]{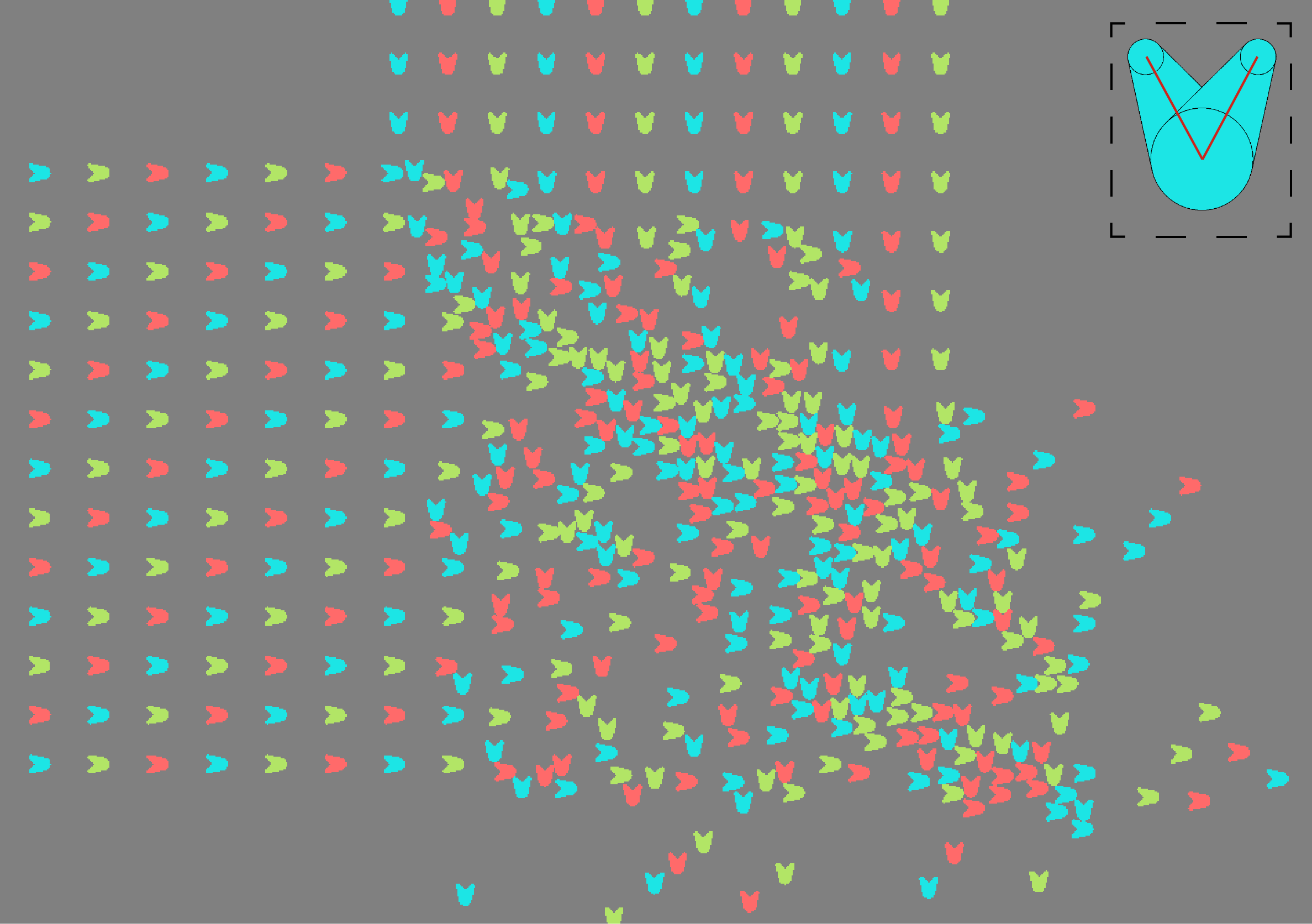}}
\subfigure[]{
\label{fig:result_3}
\includegraphics[width=0.47\columnwidth]{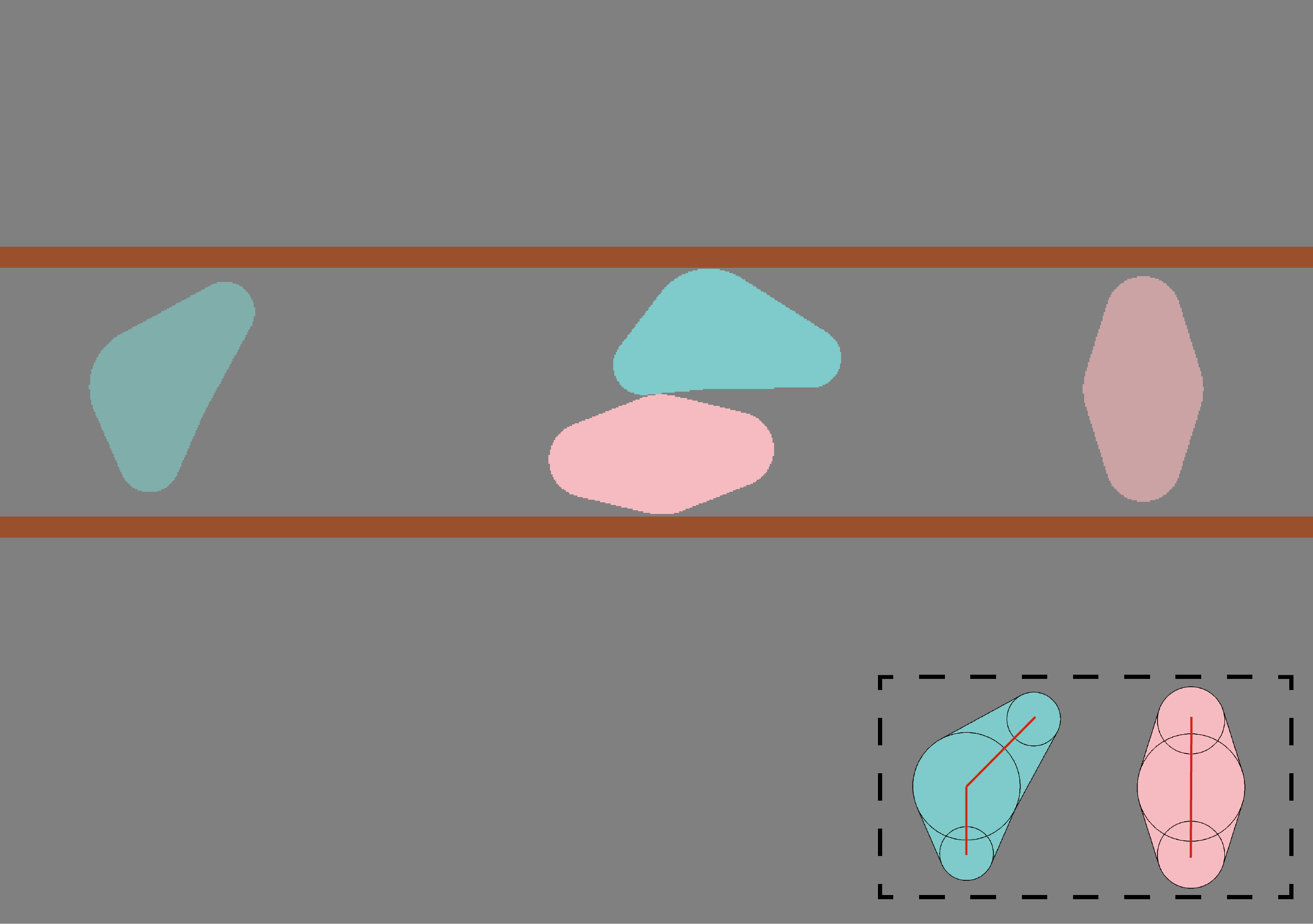}}

\subfigure[]{
\label{fig:result_4}
\includegraphics[width=0.47\columnwidth]{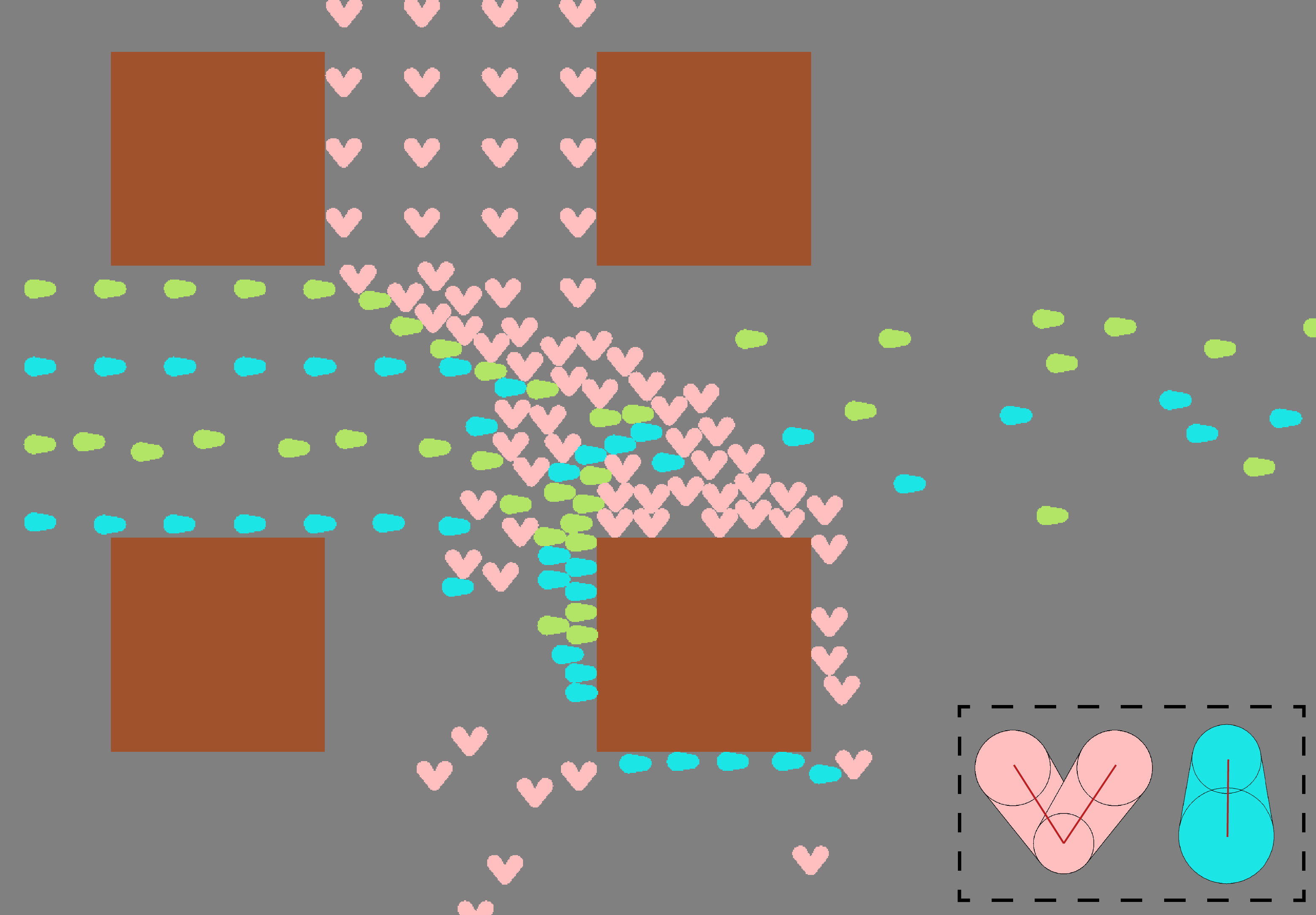}}
\subfigure[]{
\label{fig:result_5}
\includegraphics[width=0.47\columnwidth]{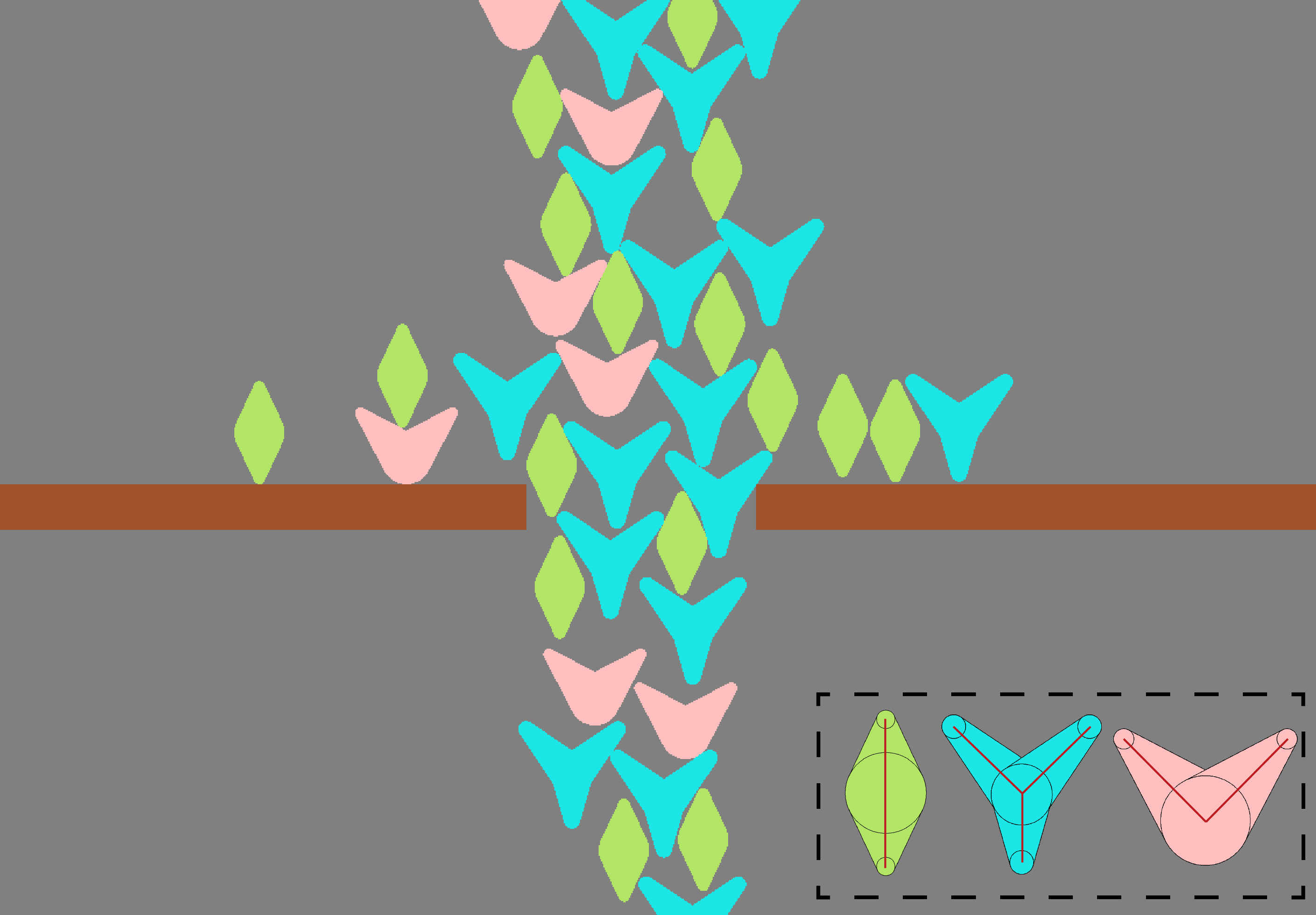}}
\caption{Different Benchmarks:
(a) $500$ agents from two vertical directions walk ahead; (b) Two agents approach and rotate in the narrow hallway; (c) $160$ agents from two vertical directions walk ahead in 4-square scene. (d) $50$ agents walk through a narrow door; CTMAT and MATRVO are able to perform collision-free navigation in these scenarios.}
\label{fig:results}
\end{figure}

In order to evalute the efficiency of our algorithm, we compare the performance of ORCA~\cite{van2011reciprocal}, ERVO~\cite{best2016real} with precomputation, MATRVO and MATRVO with precomputation in different scenarios, like  the antipodal circle. We highlight the average frame update time as a function of the number of agents in Fig. \ref{fig:comparisonOne}. Agents use one tuple for CTMAT in this comparison. We observe that MATRVO with precomputation is at most  $2$X slower than ORCA. The performance of agents with different numbers of tuples without precomputation is shown in Fig. \ref{fig:comparisonTwo}. 

During the process of computing collision-free trajectories, we need check for collisions between the agents. Representations that are too conservative may result in a high number of false positives and may not be able to compute collision-free trajectories in dense situations. Currently, we use a brute-force method to perform collision checks between the exact polygonal representations of the agents and use that data as the ground truth and compute the number of false positives for ORCA, ERVO, and MATRVO for $50$ agents in the anti-podal benchmark (see Table 1).
 
\begin{table}
\setlength{\belowcaptionskip}{10pt}%
\begin{tabular}{|p{1cm}<{\centering}|p{1cm}<{\centering}|p{1.5cm}<{\centering}|p{1.5cm}<{\centering}|p{1.5cm}<{\centering}|}
\hline
\multirow{1}*{Test} &Agent & ORCA~\cite{van2011reciprocal} & ERVO~\cite{best2016real} & MATRVO  \\
\hline
\multirow{2}*{1} &  \includegraphics[width=0.05\columnwidth]{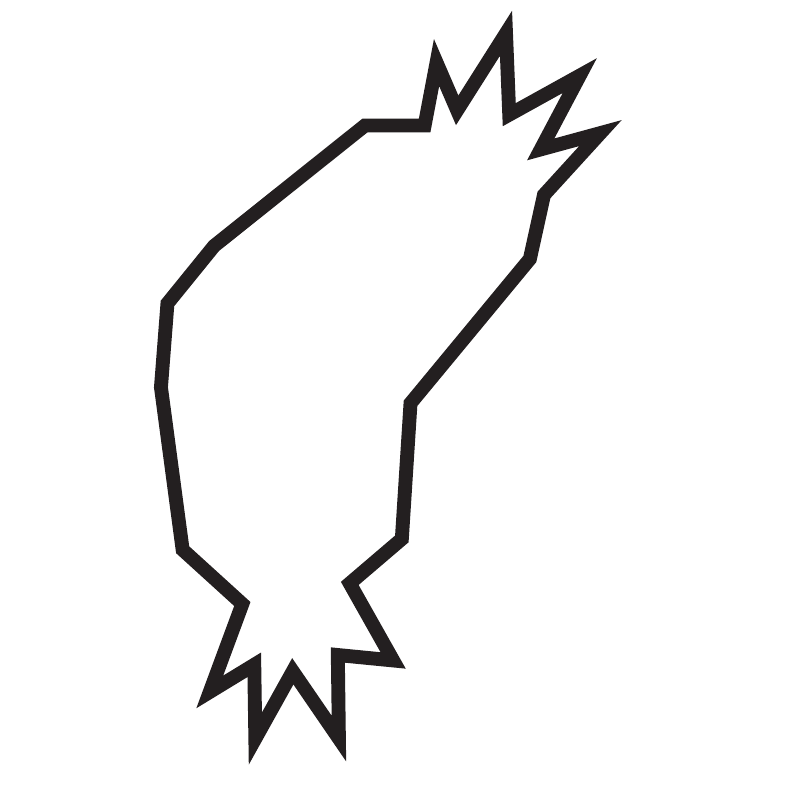} & \includegraphics[width=0.05\columnwidth]{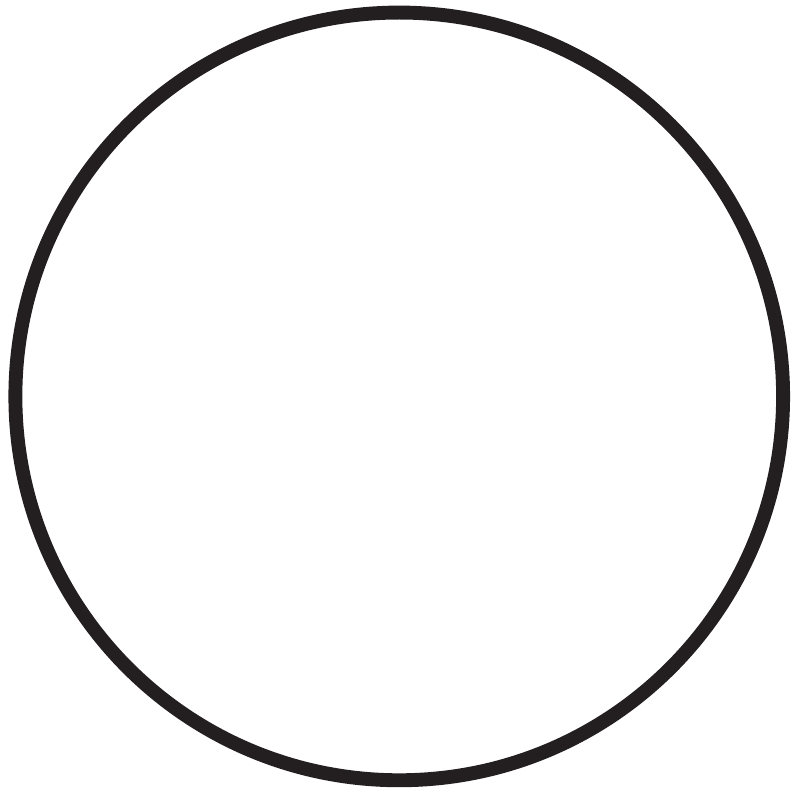} & \includegraphics[width=0.05\columnwidth]{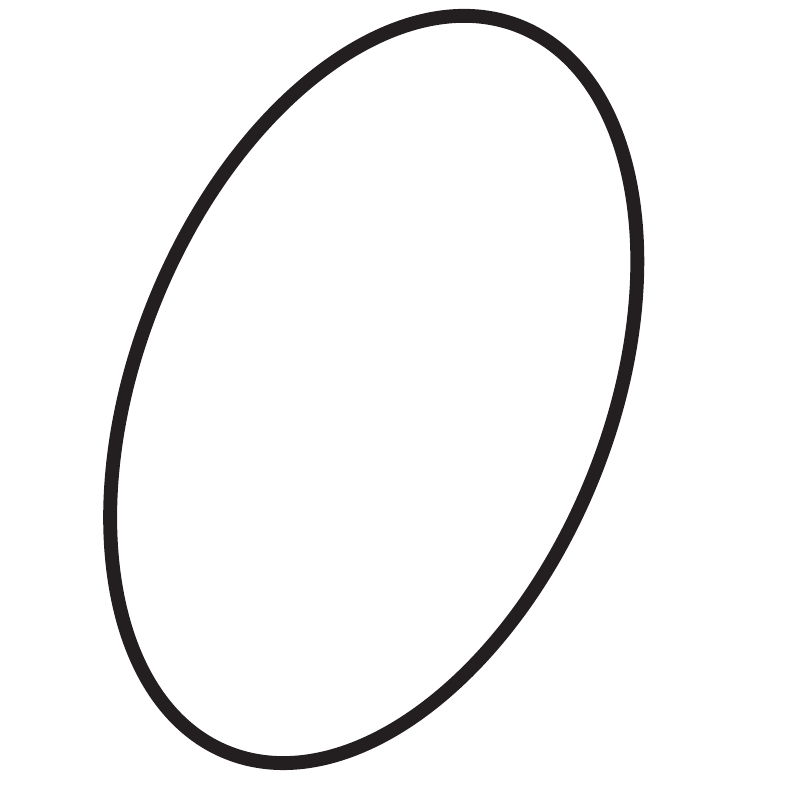} & \includegraphics[width=0.06\columnwidth]{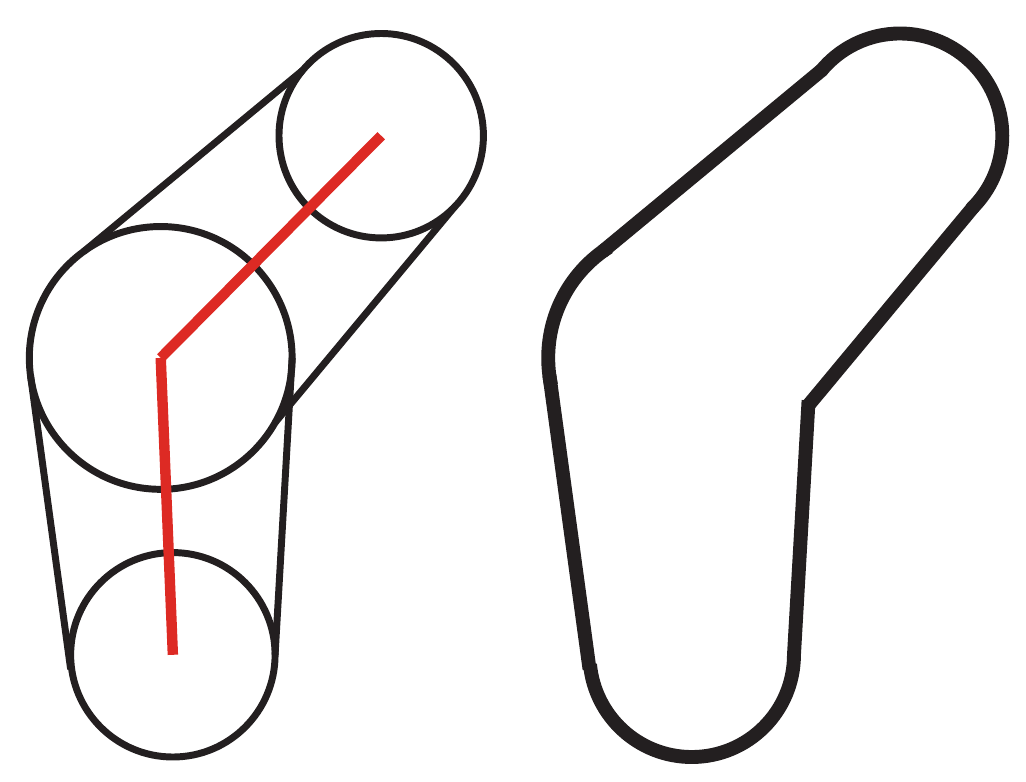} \\
\cline{3-5}
                     &&54.5\%   & 37.9\% & 8.5\%  \\
\hline
\multirow{2}*{2} &  \includegraphics[width=0.05\columnwidth]{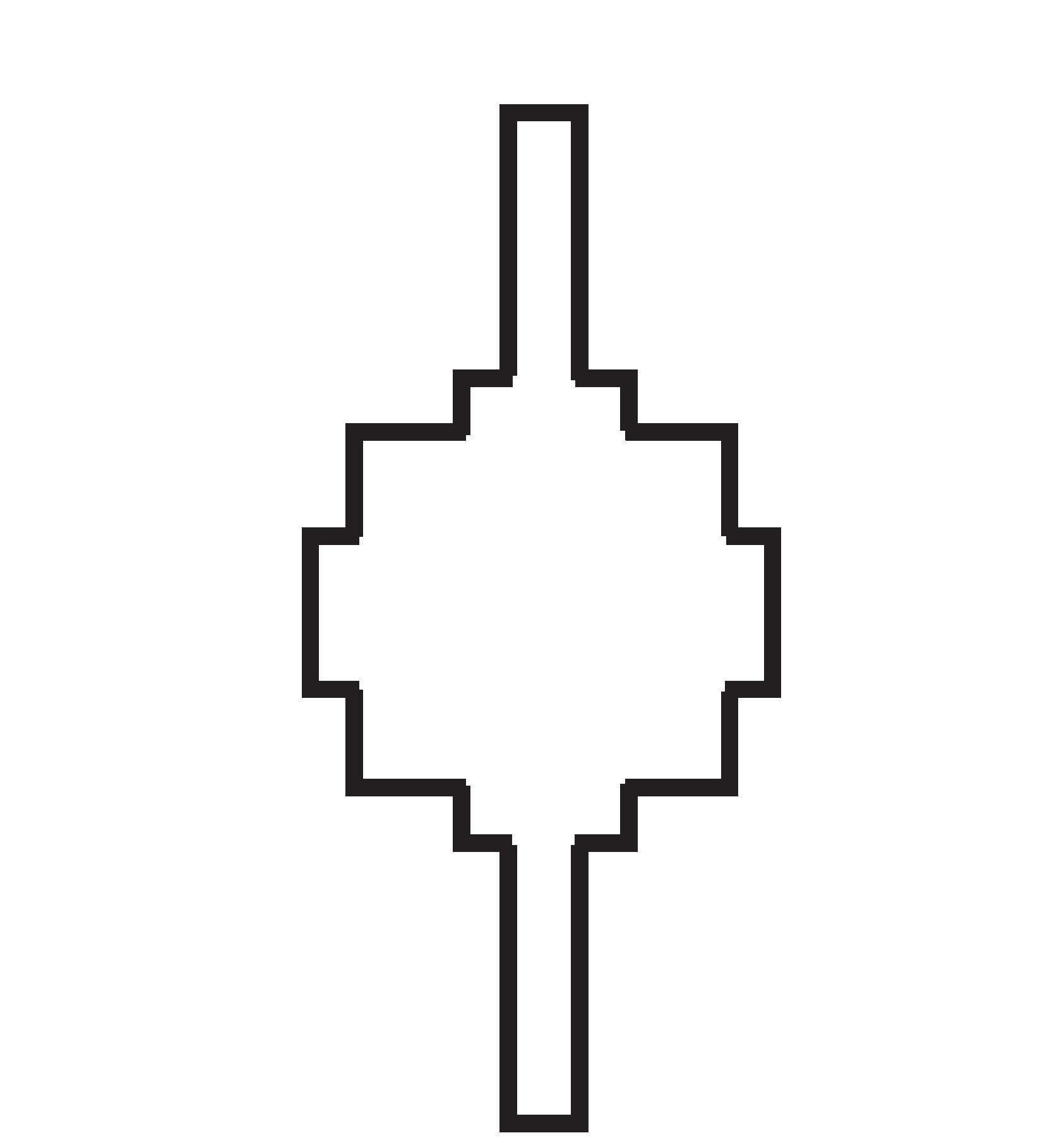} & \includegraphics[width=0.05\columnwidth]{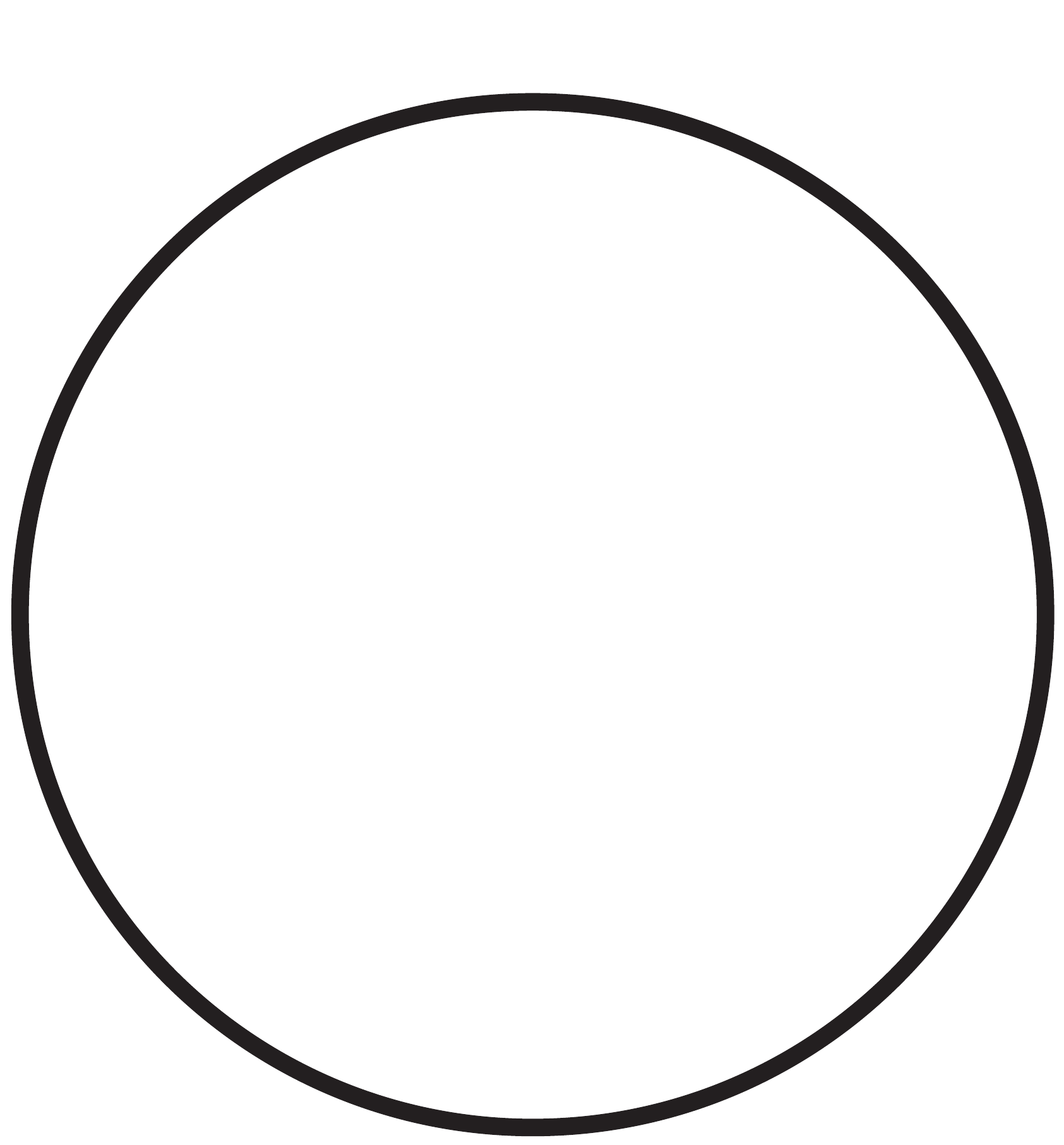} & \includegraphics[width=0.05\columnwidth]{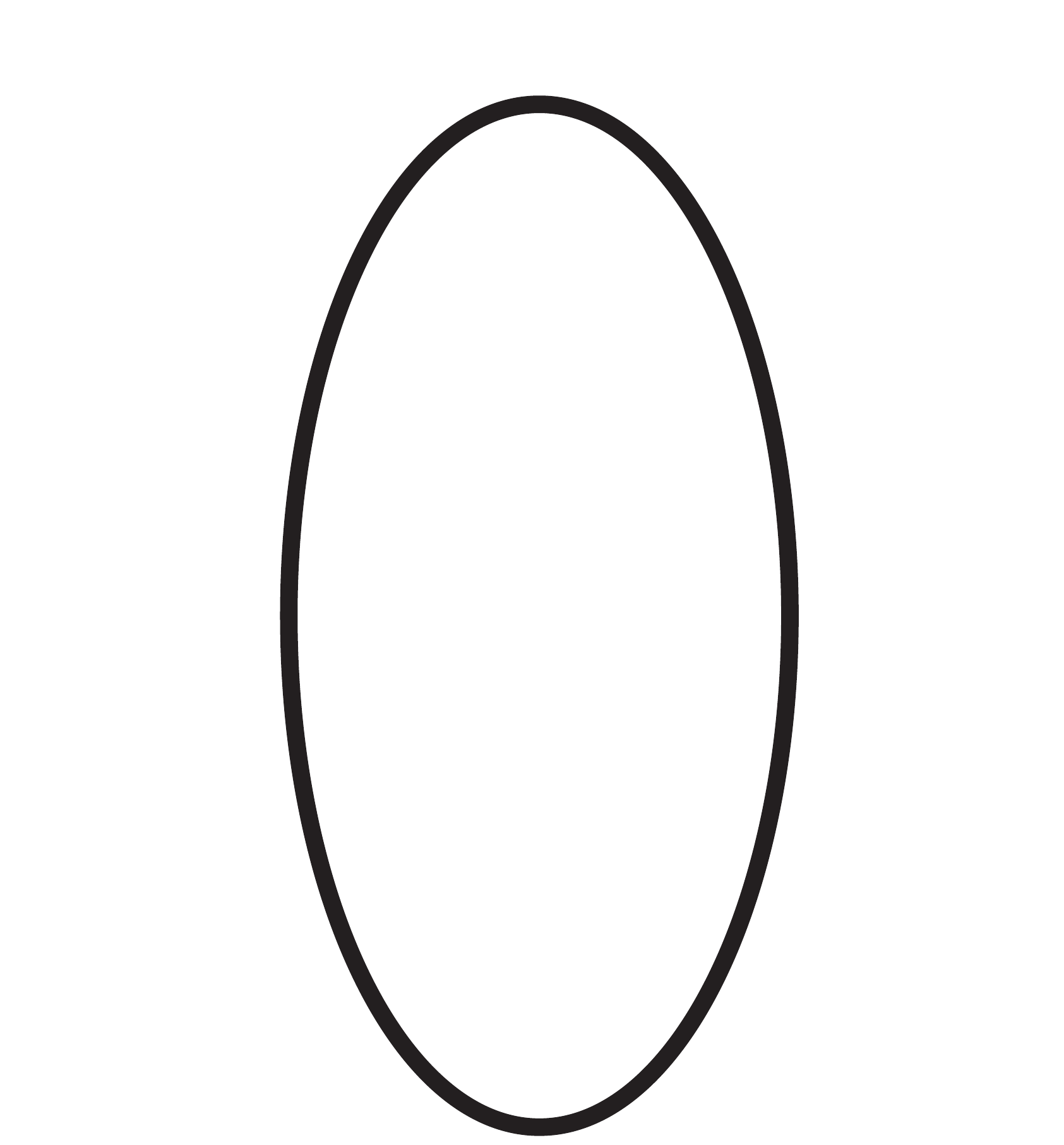} & \includegraphics[width=0.06\columnwidth]{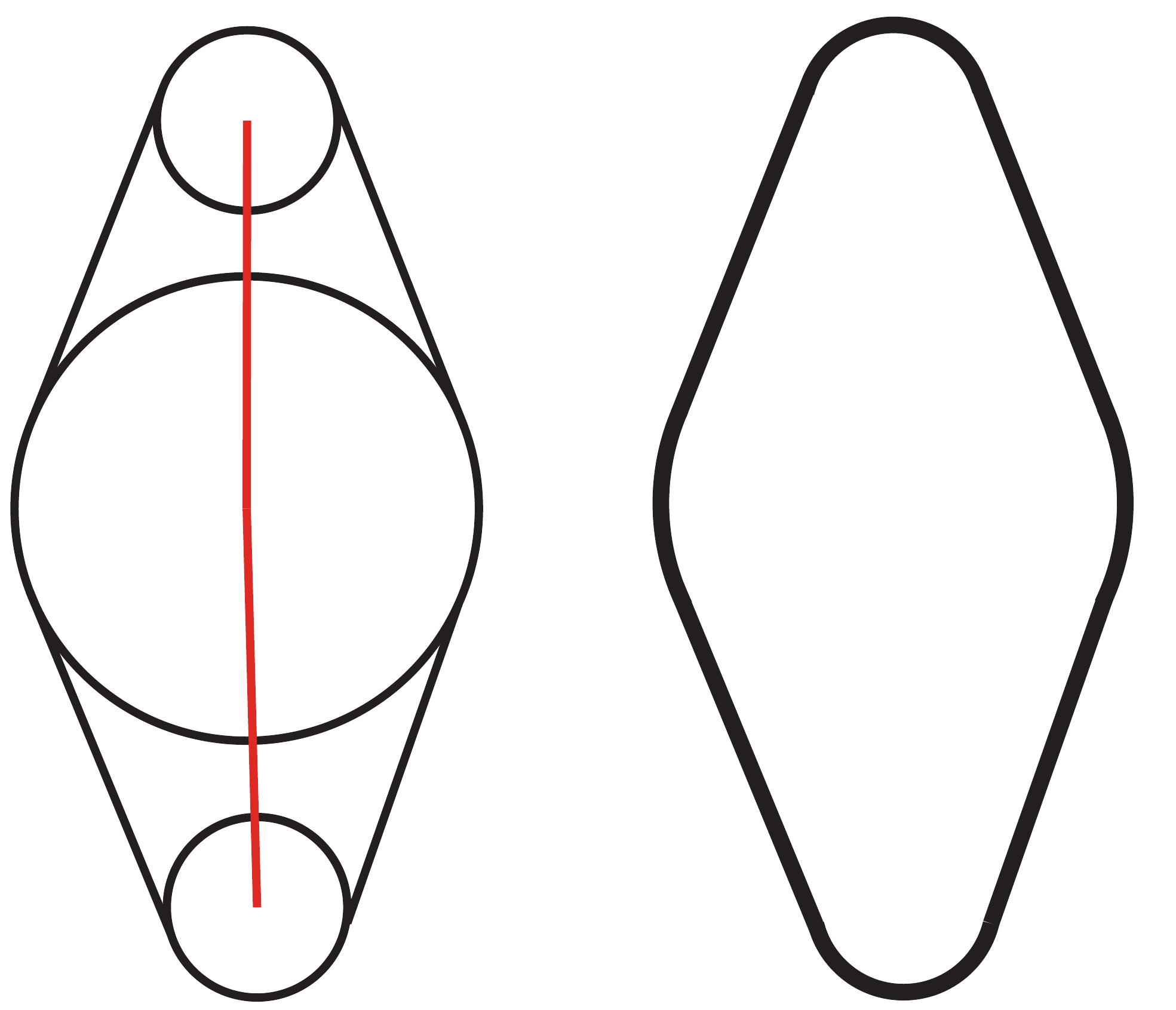} \\
\cline{3-5}
                     && 60.4\%   & 31.4\% & 9.8\%  \\
\hline
\multirow{2}*{3} &  \includegraphics[width=0.05\columnwidth]{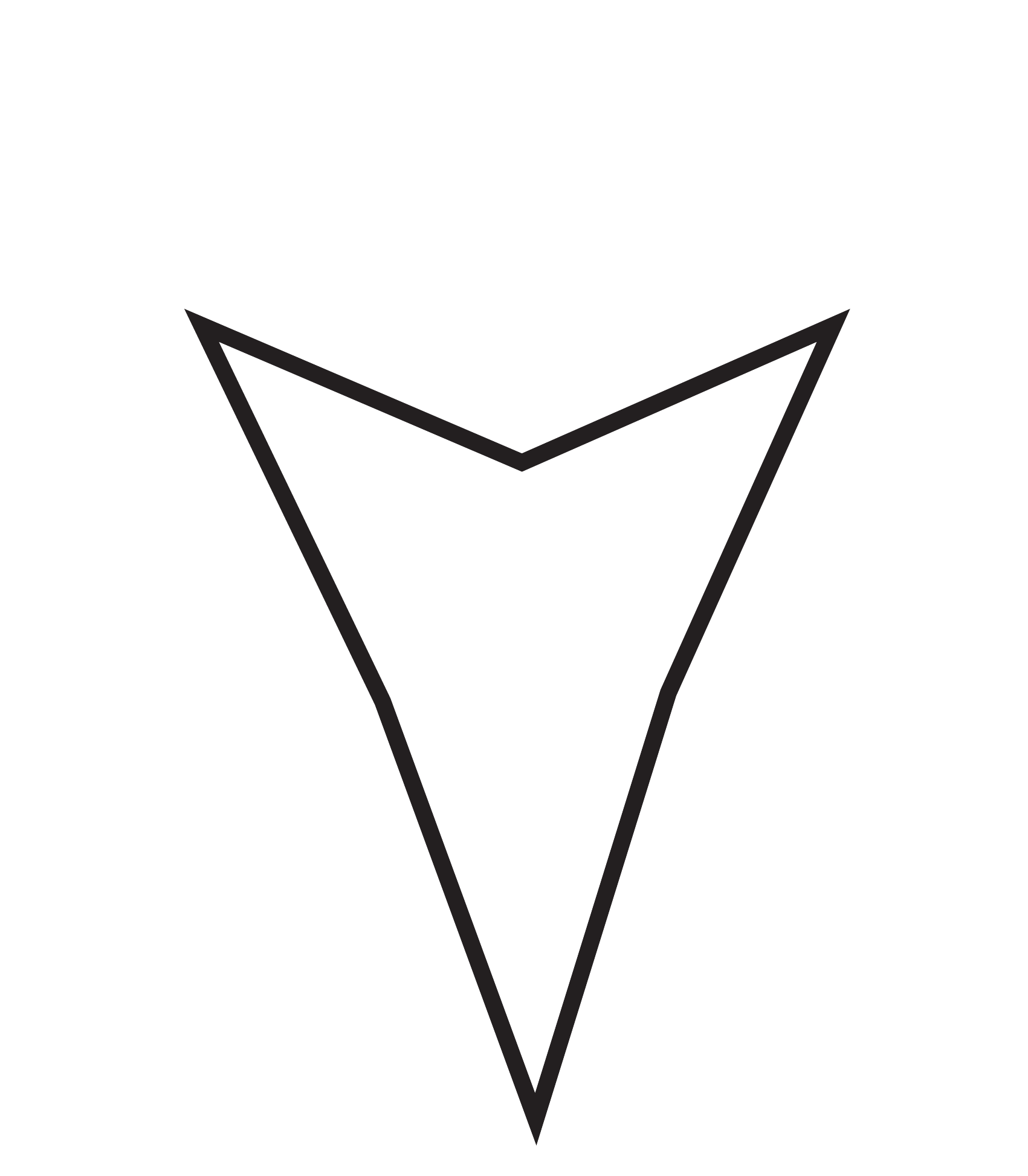} & \includegraphics[width=0.05\columnwidth]{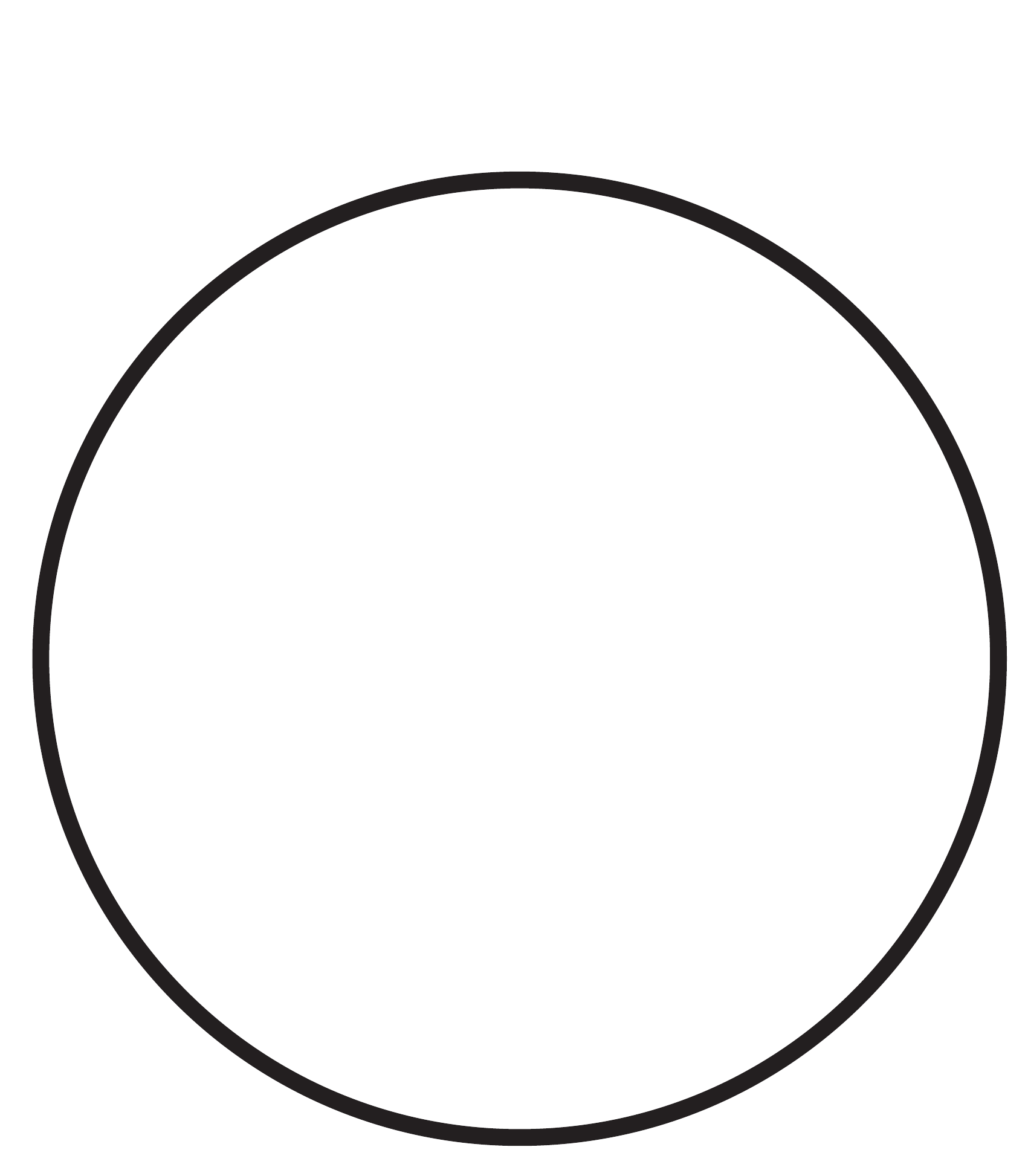} & \includegraphics[width=0.05\columnwidth]{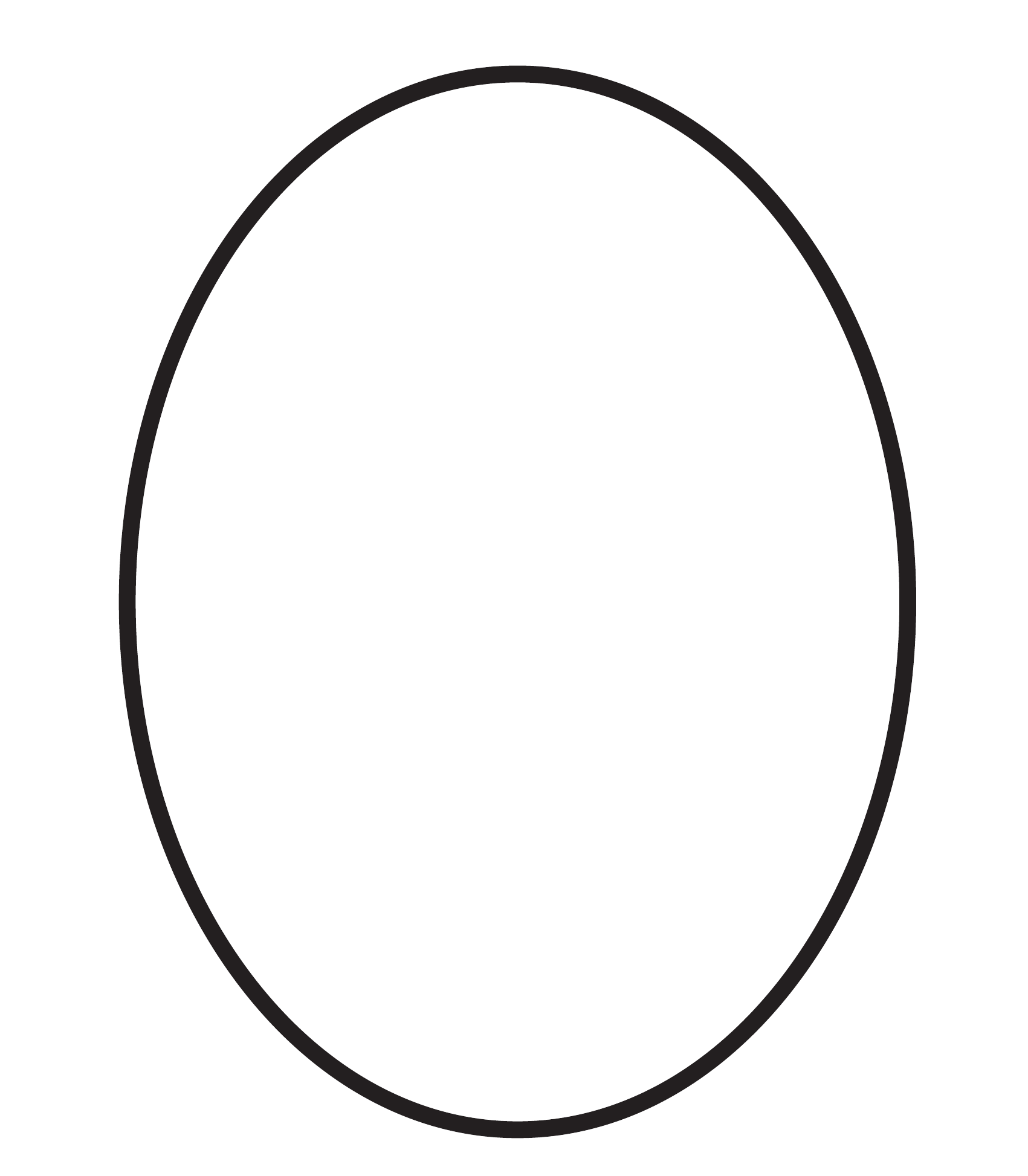} & \includegraphics[width=0.07\columnwidth]{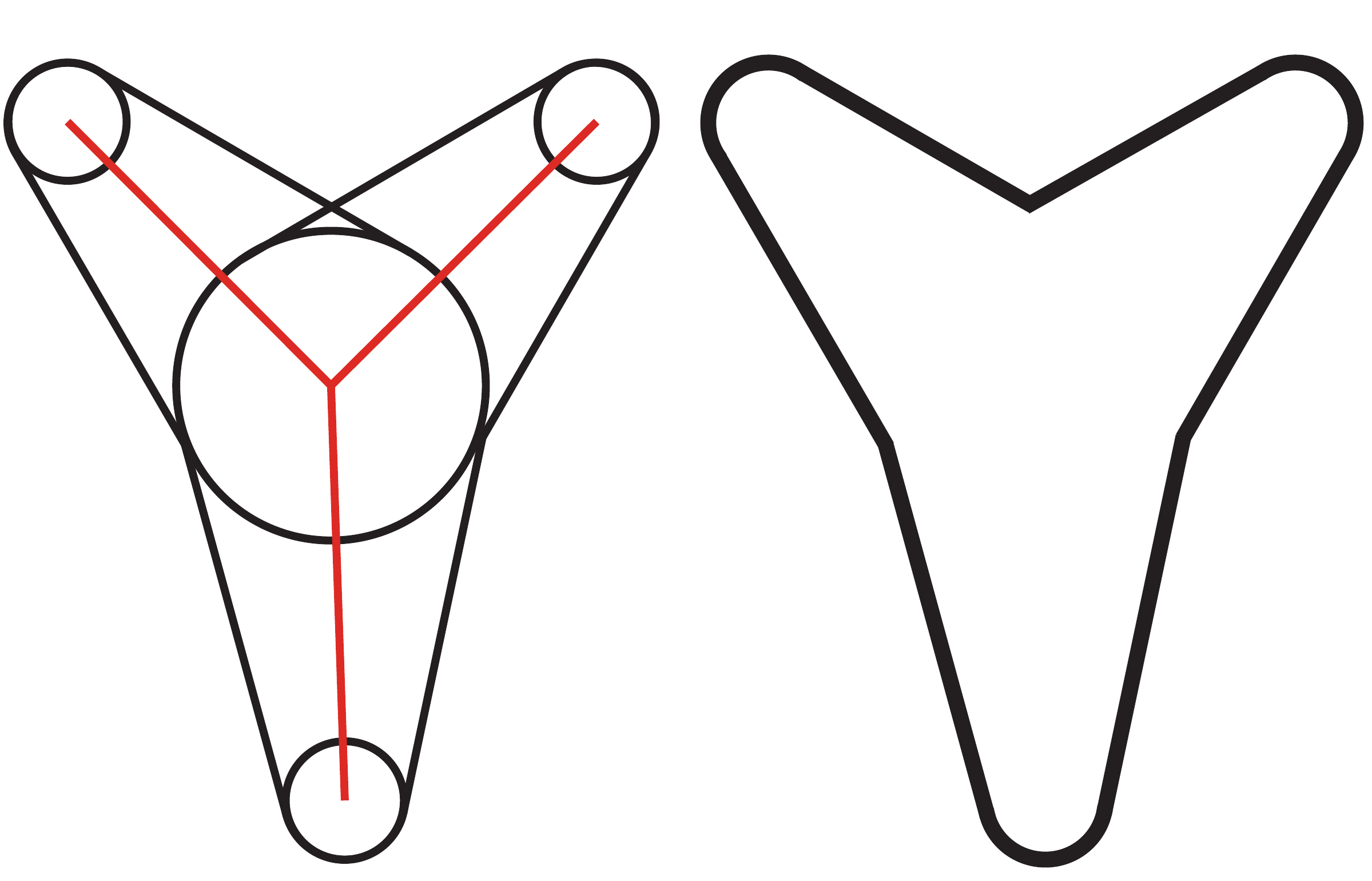} \\
\cline{3-5}
                     && 47.3\%   & 46.2\% & 9.2\%  \\
\hline
\end{tabular}
\label{table:comparison}
\caption{Comparison of ratios of false positives of ORCA, ERVO, and MATRVO in antipodal circle scenario.}
\vspace*{-3ex}
\end{table}

\section{Conclusion, Limitations and Future Work}
We present a novel algorithm for reciprocal collision avoidance between heterogeneous agents. For an arbitrary-shaped agent, we represent it with CTMAT and use MATRVO to compute collision-free trajectories for multiple agents. Taking advantage of the geometrical properties of MAT, our representation is less conservative and more flexible than current disc or ellipse-based approximation. Moreover, we can handle both convex or non-convex agents. Due to the simplicity of the formulation, MATRVO is very fast and can be used for interactive multi-agent navigation of thousands of agents on a single CPU core. We demonstrate the performance of our algorithm in simulating different scenarios and highlight the benefits over prior multi-agent navigation schemes. 

Our approach has some limitations. First, the new velocity and agent's orientation are not computed simultaneously. Second, like other VO-based methods, our algorithm also assumes perfect sensing and does not take into account uncertainty. In real traffic, different kinds of vehicles have different dynamics and we need to take them into account. As part of future work, we would like to overcome these limitations and evaluate the performance of MATRVO in more complex scenarios.

\section{Acknowledgements}
This work was supported by Hong Kong RGC Grant (HKU 717813E). We also wish to
thank Xinge Zhu for his valuable suggestions.


\bibliographystyle{ACM-Reference-Format}  
\bibliography{sample-bibliography}  

\end{document}